\documentclass[11pt,a4paper]{article}
\usepackage[hyperref]{modified_emnlp}
\usepackage{times}
\usepackage{latexsym}
\usepackage{amssymb}

\usepackage{booktabs}
\usepackage{array,multirow}

\usepackage{microtype}
\usepackage{tabularx}
\usepackage{graphicx}
\usepackage{subcaption}

\aclfinalcopy 

\usepackage{xcolor}
\definecolor{cadet}{rgb}{0.33, 0.41, 0.47}
\definecolor{carmine}{rgb}{0.59, 0.0, 0.09}

\usepackage{amsmath}

\title{Static and Animated 3D Scene Generation from Free-form Text Descriptions}

\author{Faria Huq \\
  Bangladesh University of \\ Engineering \& Technology
  \\\And
  Nafees Ahmed \\
  Stony Brook University \\
  \\\And
  Anindya Iqbal \\
  Bangladesh University of \\ Engineering \& Technology\\
}
\begin{document}
\maketitle
\begin{abstract}
% We introduce a novel task of holistic i\textbf{mag}e and v\textbf{i}deo generation from text using s\textbf{cript} (Magicript). Our proposed two-stage framework shows promising result while reducing the problem of automated generation in a concise size and ensuring high quality output. We show how our framework provides an intuitive understanding of our model and finer control on final output over existing mechanism. Our oracle pipeline works on both images and videos.

% Dataset generation code and pretrained model available at - \url{https://github.com/ShapeGen/Holistic_Generation}
Generating coherent and useful image/video scenes from a free-form textual description is technically a very difficult problem to handle. Textual description of the same scene can vary greatly from person to person, or sometimes even for the same person from time to time. As the choice of words and syntax vary while preparing a textual description, it is challenging for the system to reliably produce a consistently desirable output from different forms of language input. The prior works of scene generation have been mostly confined to rigorous sentence structures of text input which restrict the freedom of users to write description. In our work, we study a new pipeline that aims to generate static as well as animated 3D scenes from different types of free-form textual scene description without any major restriction. In particular, to keep our study practical and tractable, we focus on a small subspace of all possible 3D scenes, containing various combinations of cube, cylinder and sphere. We design a two-stage pipeline. In the first stage, we encode the free-form text using an encoder-decoder neural architecture. In the second stage, we generate a 3D scene based on the generated encoding. Our neural architecture exploits state-of-the-art language model as encoder to leverage rich contextual encoding and a new multi-head decoder to predict multiple features of an object in the scene simultaneously. For our experiments, we generate a large synthetic data-set which contains 13,00,000 and 14,00,000 samples of unique static and animated scene descriptions, respectively. We achieve 98.427\% accuracy on test data set in detecting the 3D objects features successfully. Our work shows a proof of concept of one approach towards solving the problem, and we believe with enough training data, the same pipeline can be expanded to handle even broader set of 3D scene generation problems.
\end{abstract}

\section{Introduction}
\label{sec:Introduction}

\begin{figure*}[ht]
    \begin{subfigure}[t]{0.35\textwidth}
        \centering
          \includegraphics[width=0.9\linewidth]{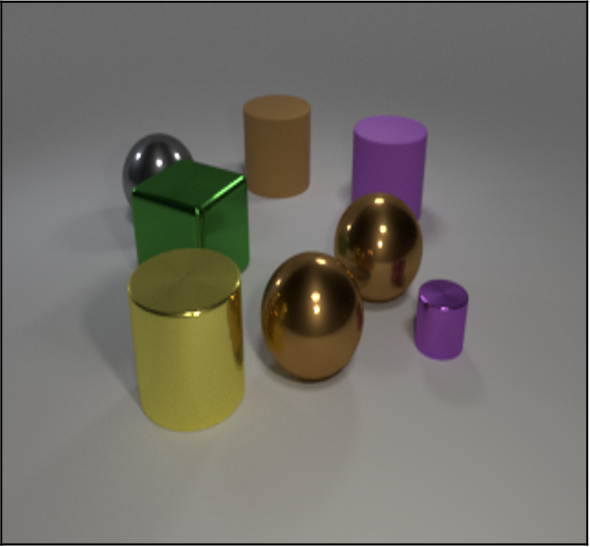}
          \caption{\textbf{Description 1:} There are one cube, three spheres and four cylinders. \\
          \textbf{Description 2:} There are two large brown spheres, a shiny yellow cylinder, a large green cube, a shiny gray sphere, a brown matte cylinder and two purple cylinders. \\
          \textbf{Description 3:} There are two spheres of brown color, a cylinder of yellow color, a cube of green color, a shiny sphere of gray color, a brown cylinder of matte texture and two cylinders of purple color.}
          \label{fig:scene_desc}
    \end{subfigure}
    \hspace{2mm}
    \begin{subfigure}[t]{0.6\textwidth}
        \centering
          \vspace{-40mm}
          
          \includegraphics[width=\linewidth]{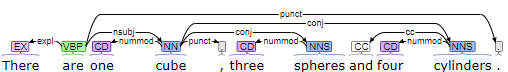}
          \caption{Dependency graph for description 1}
          \vspace{10mm}
          \label{fig:dependecy_data_1}
          
          \includegraphics[width=\linewidth]{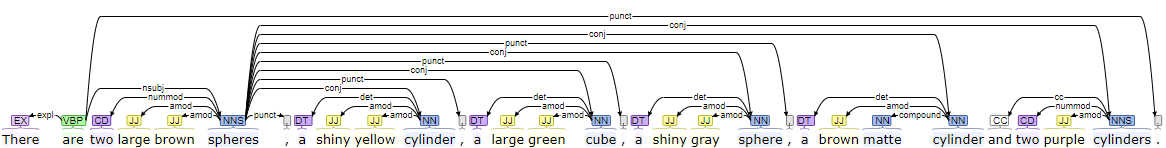}
          \caption{Dependency graph for description 2}
          \vspace{10mm}
          \label{fig:dependecy_data_2}
          
          \includegraphics[width=\linewidth]{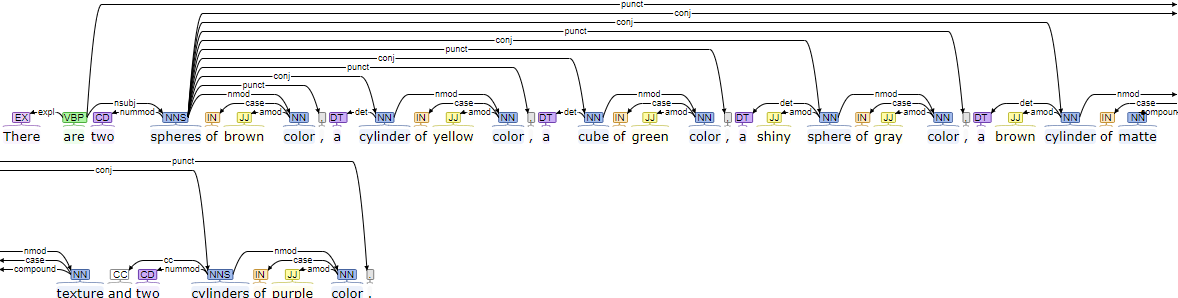}
          \caption{Dependency graph for description 3}
          \label{fig:dependecy_data_3}
    \end{subfigure}
\vspace{3mm}
\caption{A 3D scene and three of the many possible ways to describe the scene in natural language textual description (Figure \ref{fig:scene_desc}). Figure \ref{fig:dependecy_data_1}, \ref{fig:dependecy_data_2}, \ref{fig:dependecy_data_3} show the dependency graphs of each of the description shown in \ref{fig:scene_desc} (generated using \cite{parser}). As we can see, each graph is different from each other and needs different conditions to get information about the objects. Hence, we will need a large number of graphs with their associated rules to map all possible descriptions and it is not practically feasible.}
\label{fig:intro_example}
\end{figure*}

%\anindya{The first para is a bit out of context. It would be better if you can setup the context first. For example, why rendering is used, what are the limitation of current rendering processes?}
% image needs to be inserted
Natural language remains to be the most comprehensive and flexible way for human to communicate and conceptualize ideas. Although versatile, it still is imperfect and varies greatly, since every person has a different way of articulating a concept or a scene. For example, in Figure \ref{fig:scene_desc}, it is possible to describe this simple three-dimensional (3D) scene of eight basic shapes in many ways. Three possible descriptions are shown in the figure. Human beings are generally intellectually capable of interpreting the description of a 3D scene. However, automating the process using an AI (Artificial Intelligence) is not straightforward at all. It becomes even more difficult as the users vary in style of expression to depict a scene in mind. To develop an automated system that can generate a scene from textual description effectively, it needs more technical advancement in the area of Artificial General Intelligence (AGI)~\cite{AGI} and we are far behind from that. If such a text-to-scene generation system can be successfully built, it will find it's application in many domains including education, health care, animation production, game and entertainment industry, simulation and interior design, etc.
%\kuntal{You can describe here the practical benefits of solving this problem.}
%\anindya{You may introduce the challenge of moving image or others that you addressed here.}
%now to understand these descriptions,.....
% such a tool .....

Researchers have been trying to improve text-to-scene generation for a long time. The concept of scene generation from natural language description was first introduced by~\cite{adorni1983natural}. \cite{wordseye} developed a system that takes a text description and renders a scene. However, their generated scene may not always be coherent. Later, many approaches have utilized machine learning techniques to generate 3D indoor scene~\cite{chang2014learning, chang2015text, chang2017sceneseer} and infer interaction with objects~\cite{military,voxsim} from text. They convert the input description into a \textit{dependency graph}~\cite{parser} and extract the object information in a deterministic way. For example, each description shown in figure \ref{fig:scene_desc} will be mapped to different \textit{dependency graphs} (figure \ref{fig:dependecy_data_1}, \ref{fig:dependecy_data_2}, \ref{fig:dependecy_data_3}) and will need different conditions to extract the information. Hence, the users need to provide their input in a predefined structure, otherwise the input can not be processed. A pragmatic solution should be able to process \textit{free-form} text descriptions so that the users can provide their input without any constraint. It may utilize advanced deep learning techniques to design a solution for this problem as these techniques have been proved efficient to capture free-form input contexts in related fields \cite{2014sentiment,2018-review}.

\begin{figure*}[!ht]
\centering
\includegraphics[width=\linewidth]{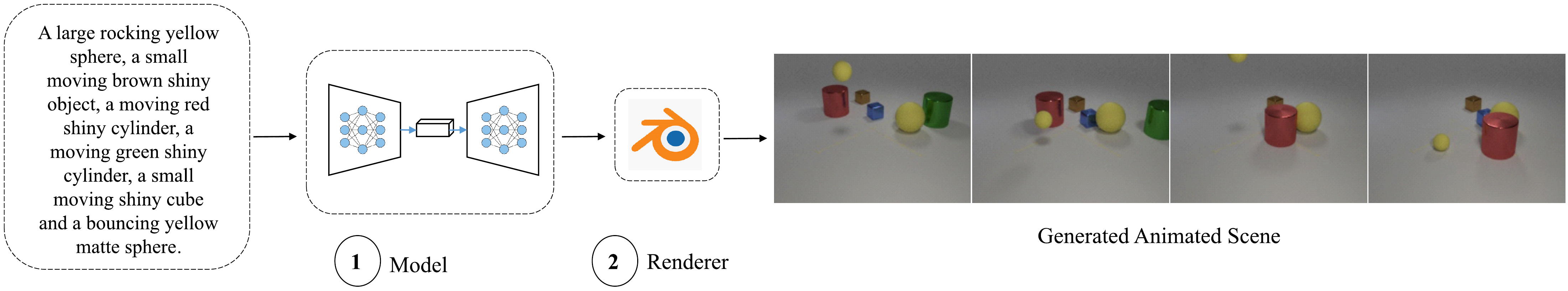}
\caption{Our proposed two-stage pipeline for text-to-scene generation. 1) The input scene description is passed through a neural architecture and the information about each object is extracted. 2) The information is passed through a renderer (made with Blender~\cite{Blender}) and the final scene is generated by grounding to an abstract layout, choosing 3D models to instantiate and arrange.}
\label{fig:motiv}
\end{figure*}

%The prior3works of scene generation are confined to specific structures of input which restrict the freedom of users to4write description. 

In this paper, we aim to design a deep learning based pipeline that can generate both static and animated scenes from free-form text descriptions. %, unlike previous studies~\cite{chang2014learning, chang2015text, chang2017sceneseer, military,voxsim}. 
For this particular study, we focus on a potential subspace of 3D scene - containing primitive shapes (cube, cylinder and sphere) and their features (color, shape, texture, motion and size). We want to understand the performance and behavior of the architecture chosen on this synthetic problem space, so that we can build upon our knowledge and understand how to approach the more generic problems. We develop a new dataset (\textit{Integrated Static and Animated Scene Generation, IScene}) for this specific purpose by synthetically generating corresponding descriptions of the scenes belonging to the widely used dataset CLEVR \cite{clevr}. \textit{IScene} contains 13,00,000 static and 14,00,000 animated scene descriptions.
Our pipeline works in two stages as shown in Figure \ref{fig:motiv}. We adopt an encoder-decoder architecture to extract necessary information from scene description and use a \textit{renderer} (i.e, software that performs rendering) to render the corresponding scene based on the extracted information. As the encoder, we choose one of the most recent language models, TransformerXL~\cite{transformerxl}. For the decoder, we design a new \textit{multi-head decoder} that is capable of creating the \textit{abstract layout} of the corresponding scene, where each \textit{head} of this decoder predicts one of the object-specific features. Here, \textit{abstract layout} represents a structured description containing the specific features of each object present in the scene and the layout is then rendered into a scene.
By enabling the prediction of multiple features simultaneously, our multi-head decoder is playing an important role in our overall architecture to understand various types of user inputs without the need of any deterministic condition. %\anindya{You may say 1/2 line signifying the contribution of this decoder to achieve the desired goal.}

%Our empirical analysis shows that our new architecture works correctly for unseen scene descriptions and captures the scene context during testing. 
We conduct empirical analysis of how our model captures the natural language contexts and generates both static and animated scenes efficiently. We also demonstrate the ability of our approach to adapt new (unseen) 3D models during inference.

Specifically, our contributions in this work are as follows - 
\begin{itemize}
    \item We design a new two-stage pipeline that can generate static as well as animated 3D scene from free-form text descriptions. In the first stage, we extract the information using a deep learning based encoder-decoder architecture. In the second stage, we render the final scene using the extracted information. %\anindya{Think about the word "novel". Write 1 line about the two stages.}
    
    \item We design a new multi-head decoder that can extract multiple object specific features to capture the scene context efficiently. To the best of our knowledge, such a multi-head decoder has not been used before for this particular task. Our new decoder plays an important role in our overall architecture to understand various types of user input. 
    We achieve up to 98.427\% accuracy on test dataset. 
    
    \item We generate a new dataset - \textit{Integrated Static and Animated Scene Generation~(IScene)}. IScene is generated programmatically under a physics-based render engine which can be extended and scaled easily. The total unique scene descriptions in this dataset are 13,00,000 and 14,00,000 for static and animated scenes, respectively. The dataset, the code for dataset generation as well as our pipeline, and the pre-trained models are available at: \url{https://github.com/oaishi/3DScene_from_text}. 
    
\end{itemize}

\section{Related Work}
\label{sec:related}
In this section, we discuss relevant previous works.

\subsection{3D scene synthesis from text description}
The concept of scene generation from natural language description was first introduced in 1983~\cite{adorni1983natural}. Wordseye~\cite{wordseye} first implemented a pipeline which used natural description to render a static scene. However, their generated scene may not always be coherent. Their input text description needs to follow a certain pattern; otherwise the rendered scene might not be consistent and meaningful.~\cite{2006real} designed a system which allows users to generate the final scene in an interactive manner from input voice and text description, where user can choose the input scene objects.~\cite{chang2014learning, chang2015text} maps the input text description into a dependency graph~\cite{parser} and learns spatial knowledge. The dependency graph is used to find out relative positioning of the scene objects.~\cite{chang2017sceneseer} allows to edit the scene and view from a specific position using natural language description and given input scene.~\cite{pigraphs,zhao2016relationship} learns interaction with daily life objects from real-world examples and generates these interactions from given input description. Similarly,~\cite{voxsim,military} generates animated scenes on a corpus of a limited amount of motions and objects.~\cite{ma2018language} can edit an input scene as per text instruction. They learn pair and group-wise positioning and existence of scene objects for coherence.         

\subsection{3D scene synthesis from Layout}
\cite{merrell2011interactive,fisher2012example} learns relative positioning from a scene database layout and synthesizes a set of plausible new scene layouts from a given input scene. \cite{fisher2015activity} provides functional plausibility of an arrangement of objects from an input 3D scan using an activity model learned from an annotated 3D scene and model database.~\cite{scenegraphnet} proposes a neural message-passing approach to complete a given incomplete 3D scene and predict object type at a specific query location.~\cite{planit} generates completes floor graph with object instances and positioning from input empty or incomplete graph. The graph is further used to instantiate a 3D scene by iterative insertion of 3D models.~\cite{zhang2020fast} presents a framework for indoor scene synthesis, while given a room geometry and a list of objects with learnt priors.~\cite{intelligent} proposes a novel 3D-house generation architecture. Their system takes linguistic description as input and uses Graph Convolutional Network to predict the room layout. Their system also includes a generative adversarial network module which generates floor textures of each room. The textures and layout are passed to the renderer to generate the final 3D house layout.%~\cite{graph2plan} adopts a VAE-GAN~\cite{gan,vae} architecture which generates a new layout of an input scene simultaneously matching the top-view of the generated layout with that of the input layout.

\subsection{Image and video generation from text description}
Over last decade, many studies proposed novel generative models for generating image from text description using Generative Adversarial Network (GAN)~\cite{gan} and Variational Autoencoder~\cite{vae}.~\cite{stackgan,AttnGAN,reed2016generative, text2shape, attribute2image,GeneratingIF,ControllableTG} propose various kinds of novel architecture for generating images from text.~\cite{scenegraph} generate images from scene graph. Some recent studies are focusing on image generation from text in an iterative manner where one or two actors engage in giving instruction about the scene~\cite{iclevr,codraw,seqattngan,multiturnGAN,text2scene,ChatPainterIT}.~\cite{storygan} generates a series of images which form a story altogether. A related study is to generate video from story plot~\cite{imaginethis}.~\cite{tcgan,videovaegan} generate videos from caption.

% \subsection{Neural Program Synthesis}

% ~\cite{3Dshapeprogram,write_exec, scene2program} show the possibility of neural program synthesis. Instead of generating images directly, they learn to describe scene with programs, these programs are used to render the final 3D scene similar to ours. ~\cite{clevrer,infer_forvq} infer programs for visual reasoning.~\cite{clevr,clevrer,cater} use script for dataset generation for the purpose of visual reasoning.

%To the best of our knowledge, this is the first study on the inclusion of Transformer for scene generation from free-form text descriptions. Furthermore, we show how our pipeline generates both static and animated scenes which have not been studied before as our best knowledge.    
%Our novel pipeline allows us to automate image and video generation together, which is difficult for prior studies in the literature. Furthermore, Our two-staged platform allows us to control and alter the scene objects even at inference, which is more practical than prior studies.

\section{Objective}
\label{sec:task}
%\textcolor{carmine}{clarify each task}

For our task of static and animated scenes generation from free-form text descriptions, we use the dataset CLEVR (A Diagnostic Dataset for
Compositional Language and Elementary Visual Reasoning)~\cite{clevr}, extensively used in the literature. CLEVR is generated in the Blender Engine~\cite{Blender}, using random combinations of primitives (Sphere, Cube and Cylinder) and \textit{object-features} (color, shape, size and texture). CLEVR also provides \textit{Compositional Generalization Test (CoGenT)}, where the dataset is divided under two conditions (conditionA, conditionB). Based on CLEVR, We generate our Integrated Static and Animated Scene Generation (IScene) dataset (See Section~\ref{sec:dataset} for details).% \anindya{How did you modify? Do you release the modified dataset?}~\oaishi{Sir, described the details in following section and adjusted the writing. sir, is this better? sir, i released the data preparation code which will generate the dataset synthetically.}

%Their homepage provides more details on the dataset \footnote{ \url{https://cs.stanford.edu/people/jcjohns/clevr/}}.

We work on CLEVR for multiple reasons. First, the object types, object features, location and camera position of each associated scene used in CLEVR are publicly available. Hence, we can use this information as ground truth to validate our approach. Second, %the associated scenes were rendered using Blender script which is publicly available, thus can be extended for our particular needs. Third, 
CLEVR uses \textit{question template} for visual reasoning. These question templates use the ground truth information for generating different types of questions. Similarly, we can design \textit{description template} for generating different types of descriptions. Third, each object has a minimum four associated features that provide a wide range of possible combinations and possible scene descriptions.

% https://tex.stackexchange.com/a/125294
\begin{figure*}[ht]
    \begin{subfigure}[b]{0.4\textwidth}
        \centering
          \includegraphics[width=0.9\linewidth]{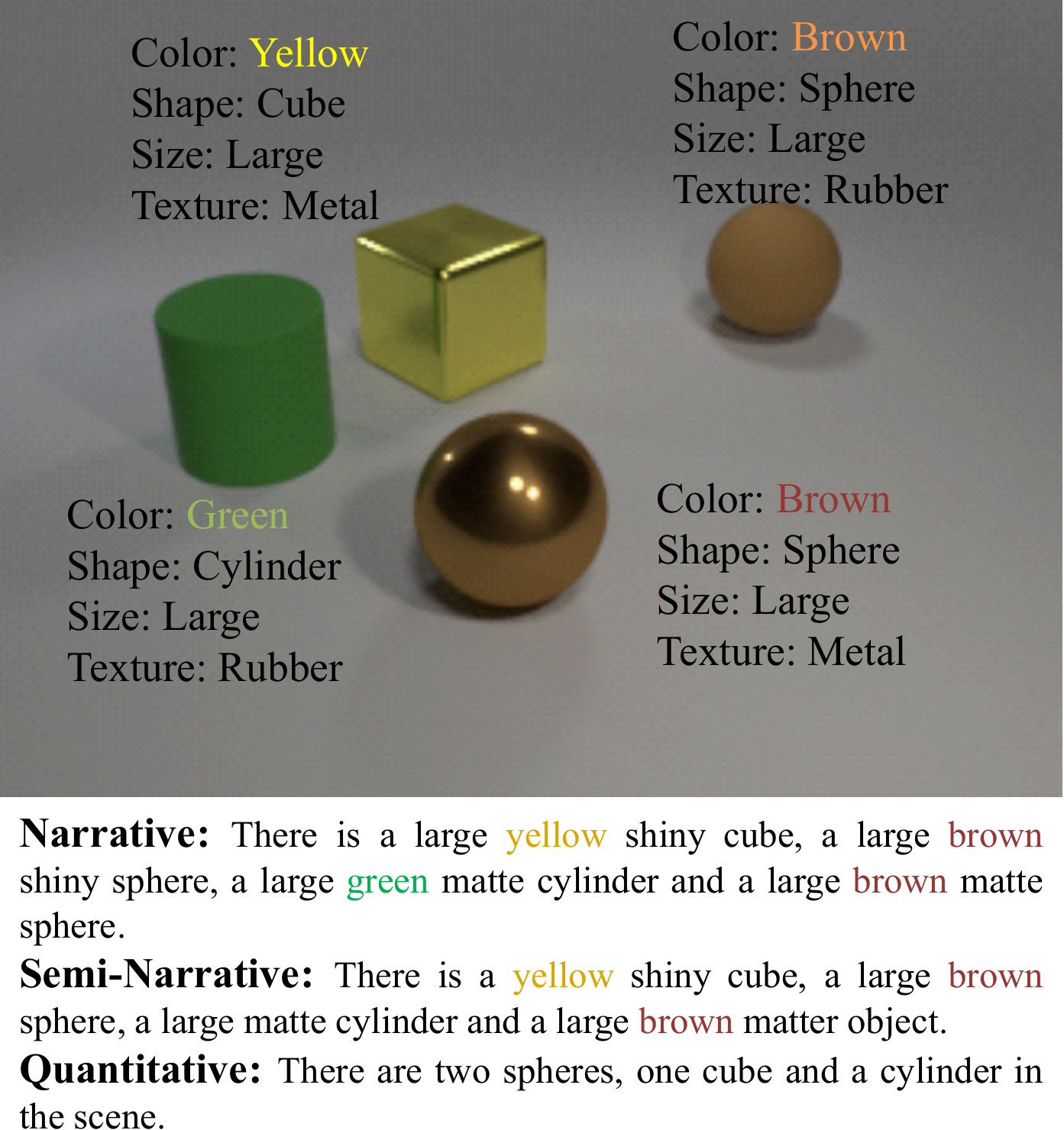}
          \caption{An example static scene with\\ corresponding description of each category}
          \label{fig:image_data}
    \end{subfigure}%
    \begin{subfigure}[b]{0.6\textwidth}
        \centering
          \includegraphics[width=\linewidth]{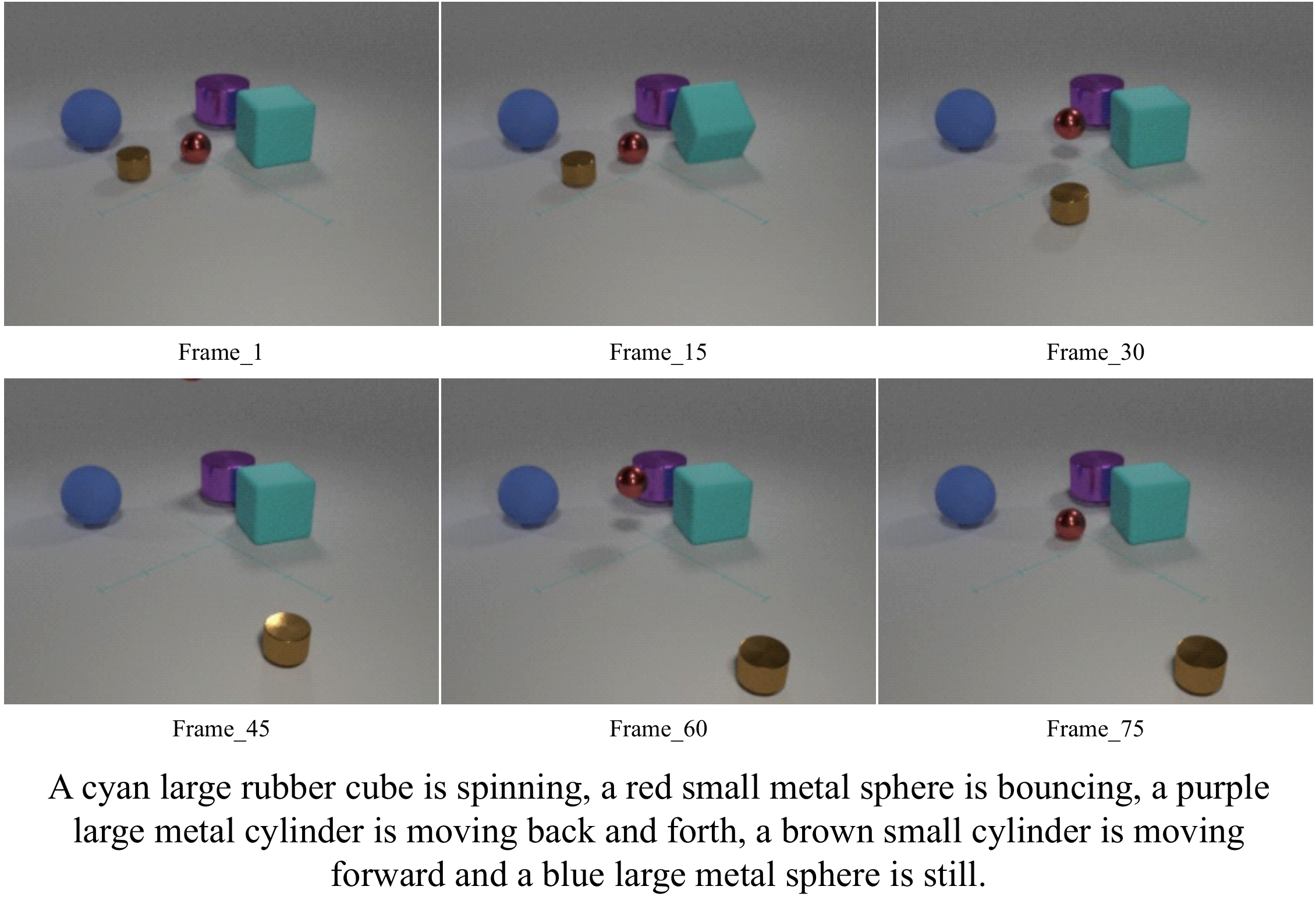}
          \caption{An example animated scene with corresponding  narrative description}
          \label{fig:video_data}
    \end{subfigure}
\vspace{2mm}
\caption{Examples from IScene Dataset} 
\end{figure*}

We will validate our approach from the following perspectives: 

1. \textbf{Ability of our approach with similar data patterns:} We will test our trained model on a similar combination of object features seen during training (referred to as conditionA/condA in CLEVR). The result will show how our pipeline captures known data patterns. We will use non-overlapping data points to train, test and validate our model.

2. \textbf{Ability of our approach with novel data patterns:} We want to test how our model performs on a novel combination of object features during inference (referred to as conditionB/condB in CLEVR). For example, any data points including \textit{red cube} is not present in the training set. During inference, we will test our model on data points containing red cube. The result will show how our model adapts to new data patterns.

3. \textbf{Editability of our approach:} We plan to test if our pipeline can provide control over edit operation. We will provide separate 3D models of the same scene objects to our renderer script and analyze if these are executed correctly.

For each of the aforementioned validation task, we consider both static and animated 3D scenes.

\section{IScene dataset}
\label{sec:dataset}
% https://tex.stackexchange.com/a/125294
\begin{figure*}[ht]
     \begin{subfigure}[b]{0.275\textwidth}
        \centering
          \includegraphics[width=\linewidth]{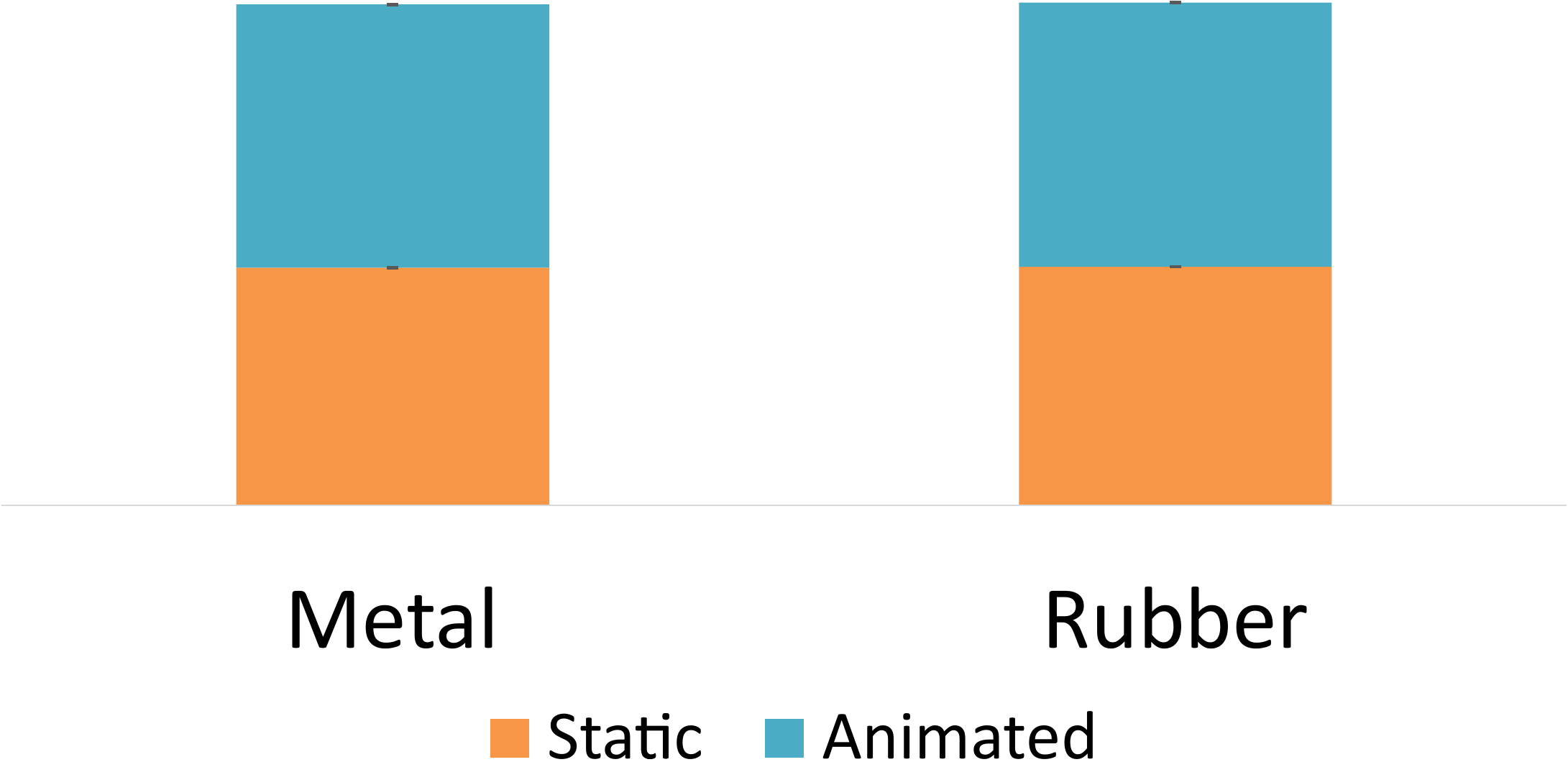}
          \caption{Distribution of texture}
          \label{fig:sfig1}
         \end{subfigure}%
         \begin{subfigure}[b]{0.275\textwidth}
        \centering
          \includegraphics[width=\linewidth]{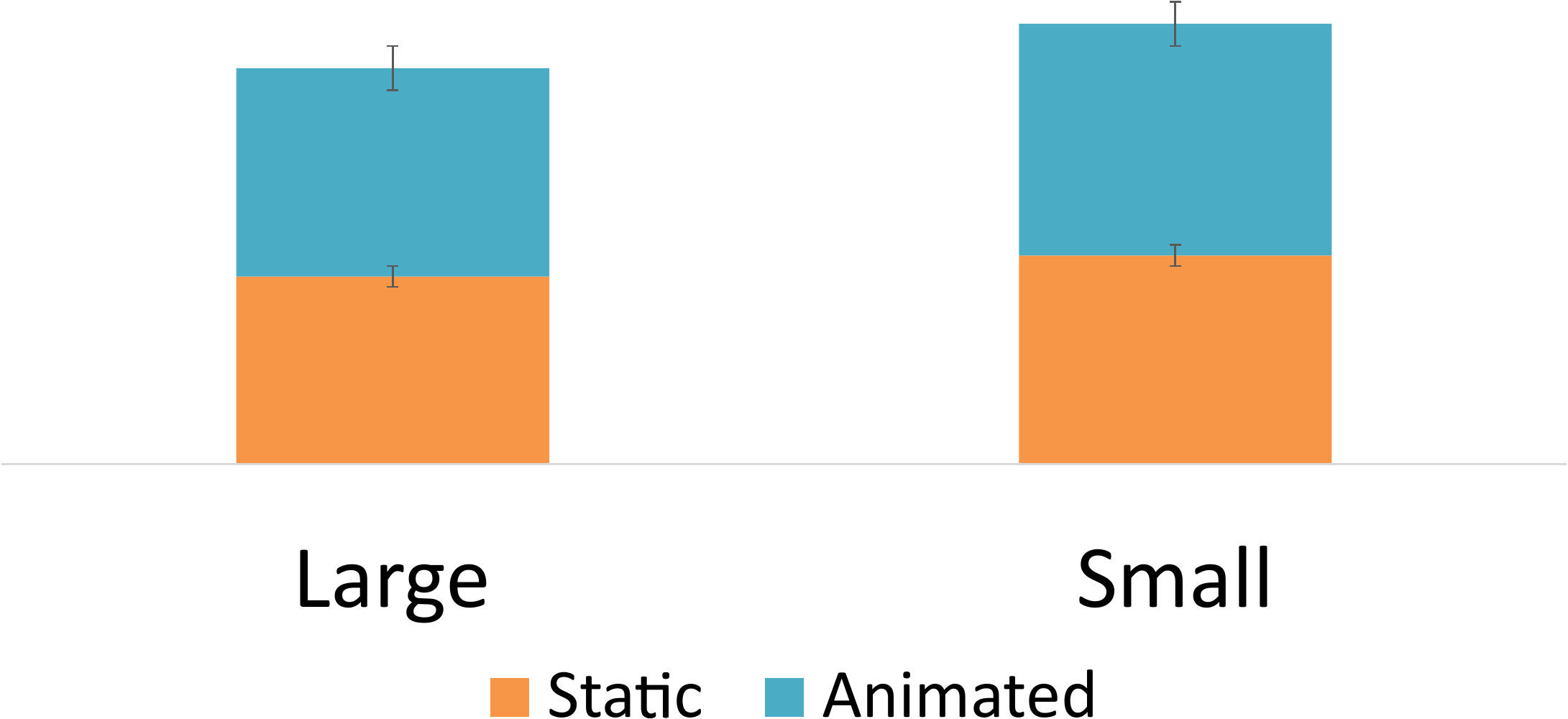}
          \caption{Distribution of size}
          \label{fig:sfig2}
         \end{subfigure}%
         \begin{subfigure}[b]{0.275\textwidth}
        \centering
          \includegraphics[width=\linewidth]{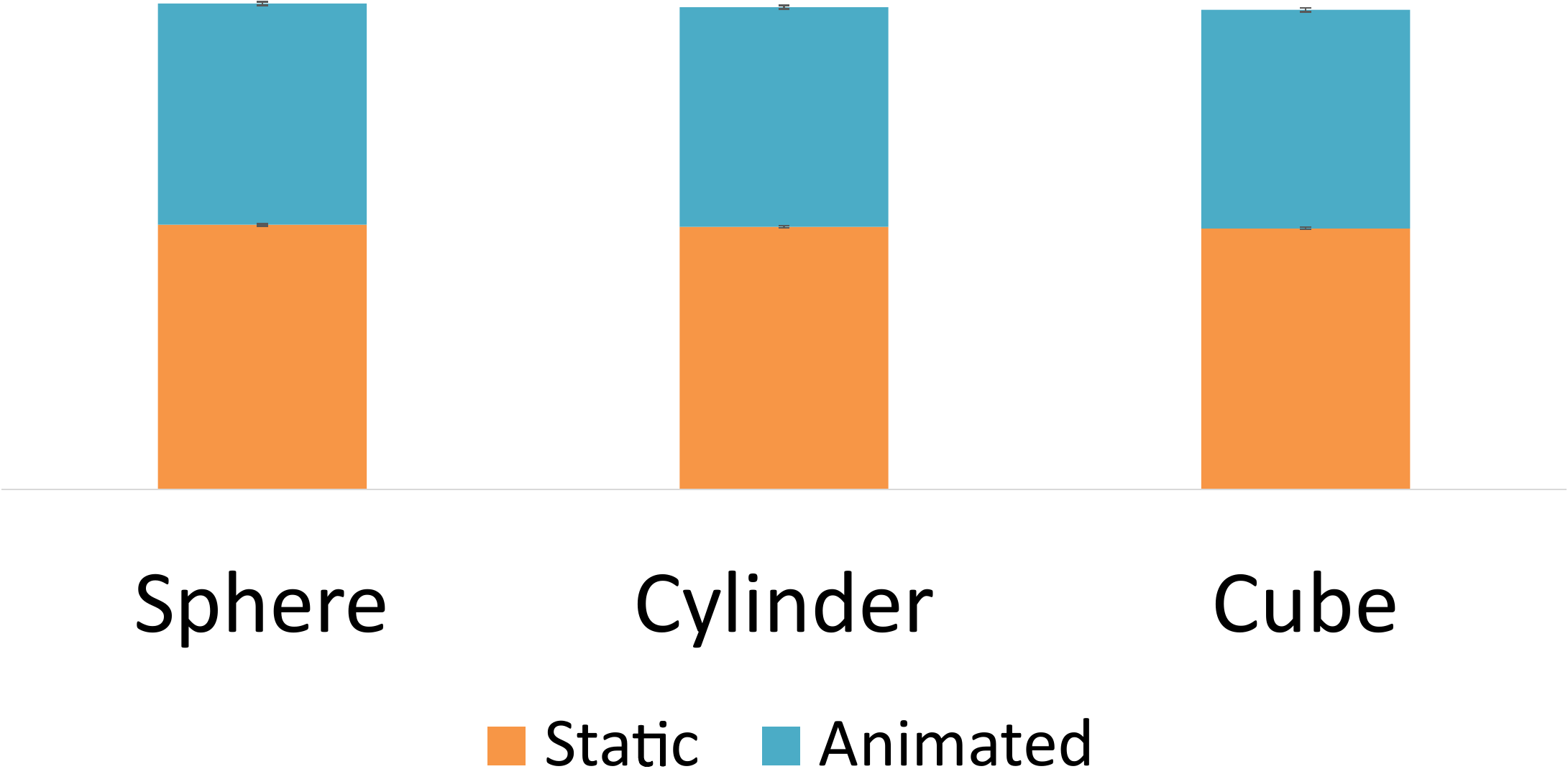}
          \caption{Distribution of shape}
          \label{fig:sfig3}
         \end{subfigure}%
         \begin{subfigure}[b]{0.15\textwidth}
        \centering
          \includegraphics[width=\linewidth]{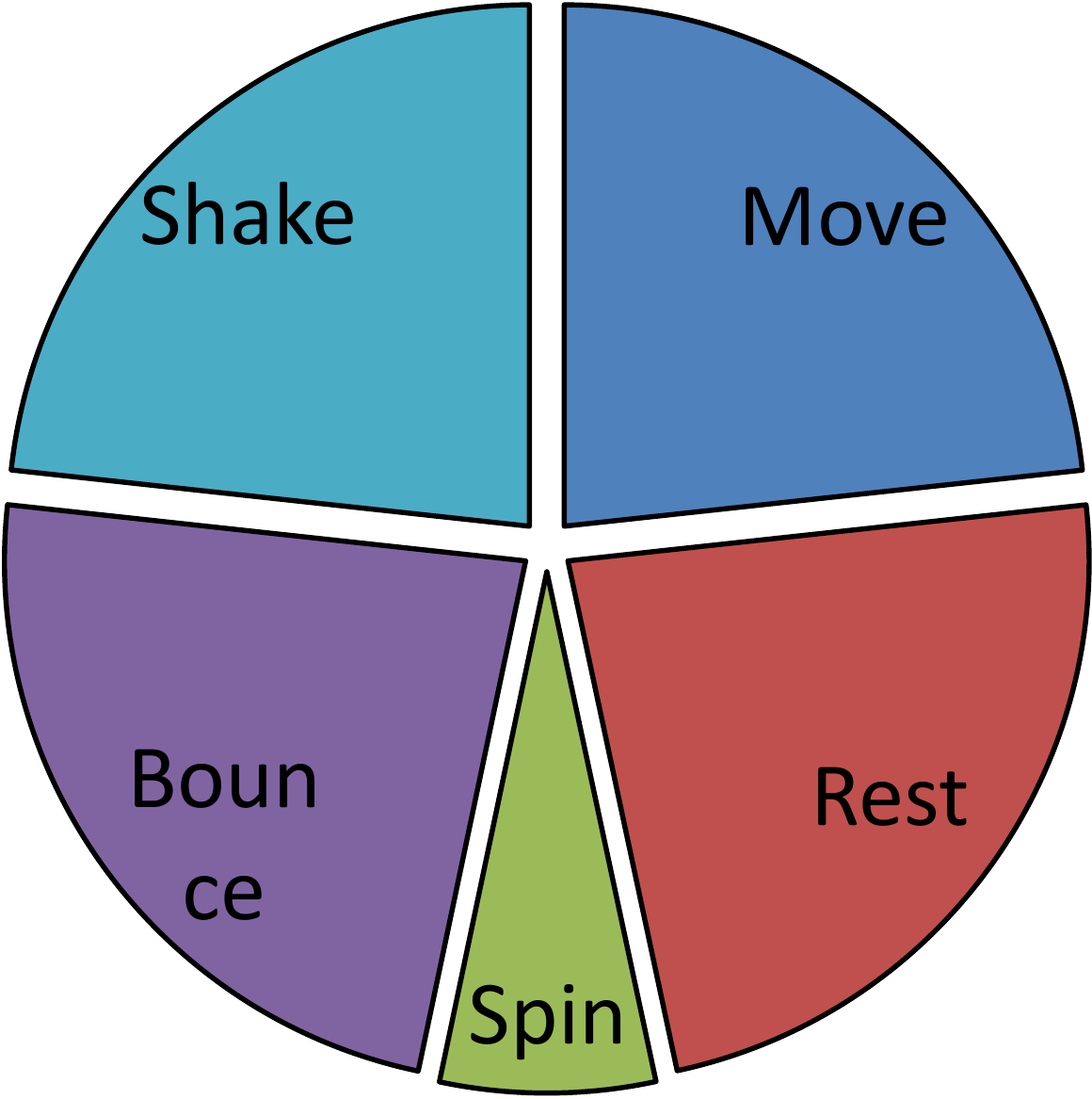}
          \caption{Distribution of motion}
          \label{fig:sfig4}
         \end{subfigure}
% leave a blank line to change row         

\vspace{6mm}

     \begin{subfigure}[b]{0.33\textwidth}
                \centering
                \includegraphics[width=\linewidth]{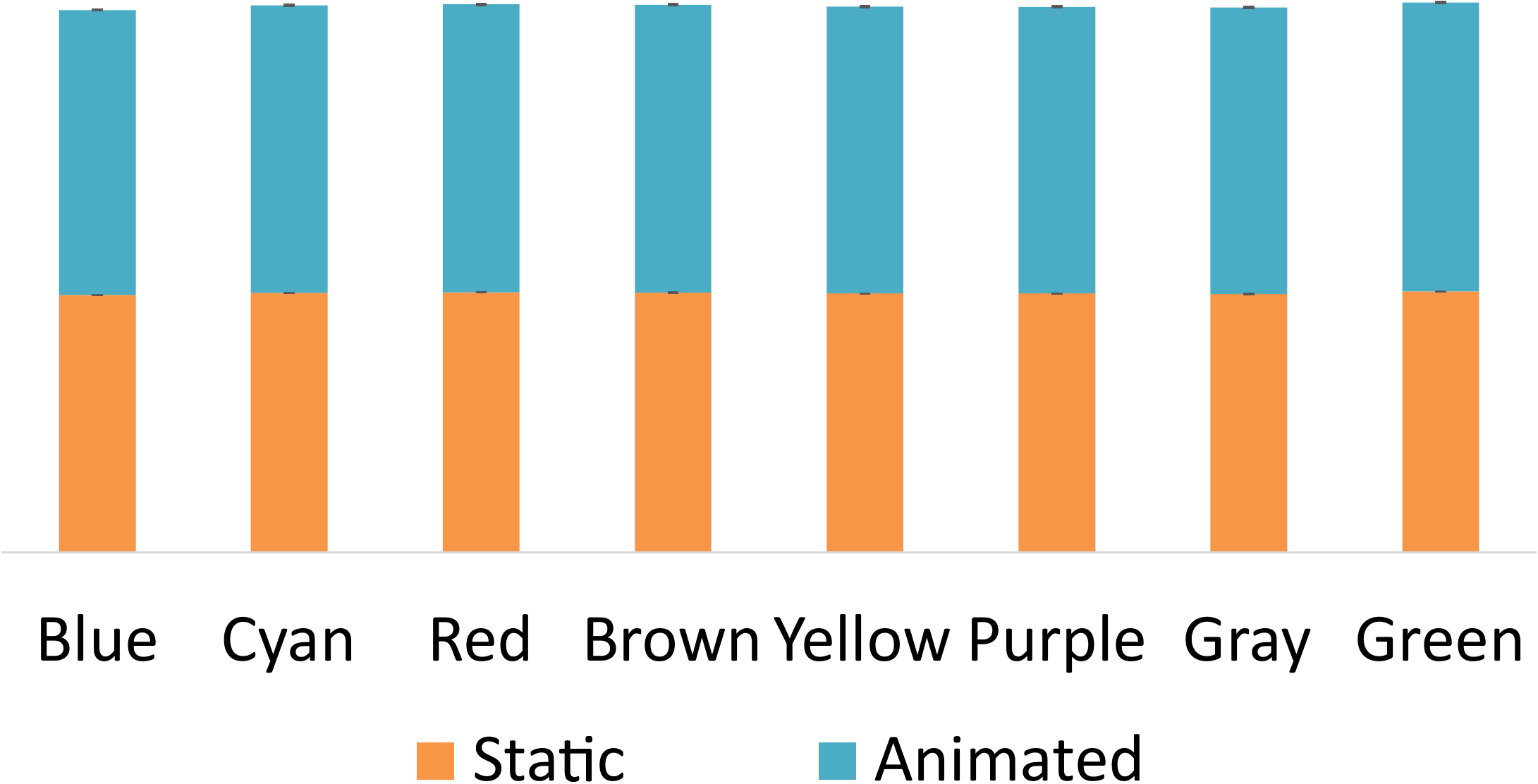}
                \caption{Distribution of color}
             \label{fig2:sfig5}
     \end{subfigure}%
     \begin{subfigure}[b]{0.33\textwidth}
                \centering
                \includegraphics[width=\linewidth]{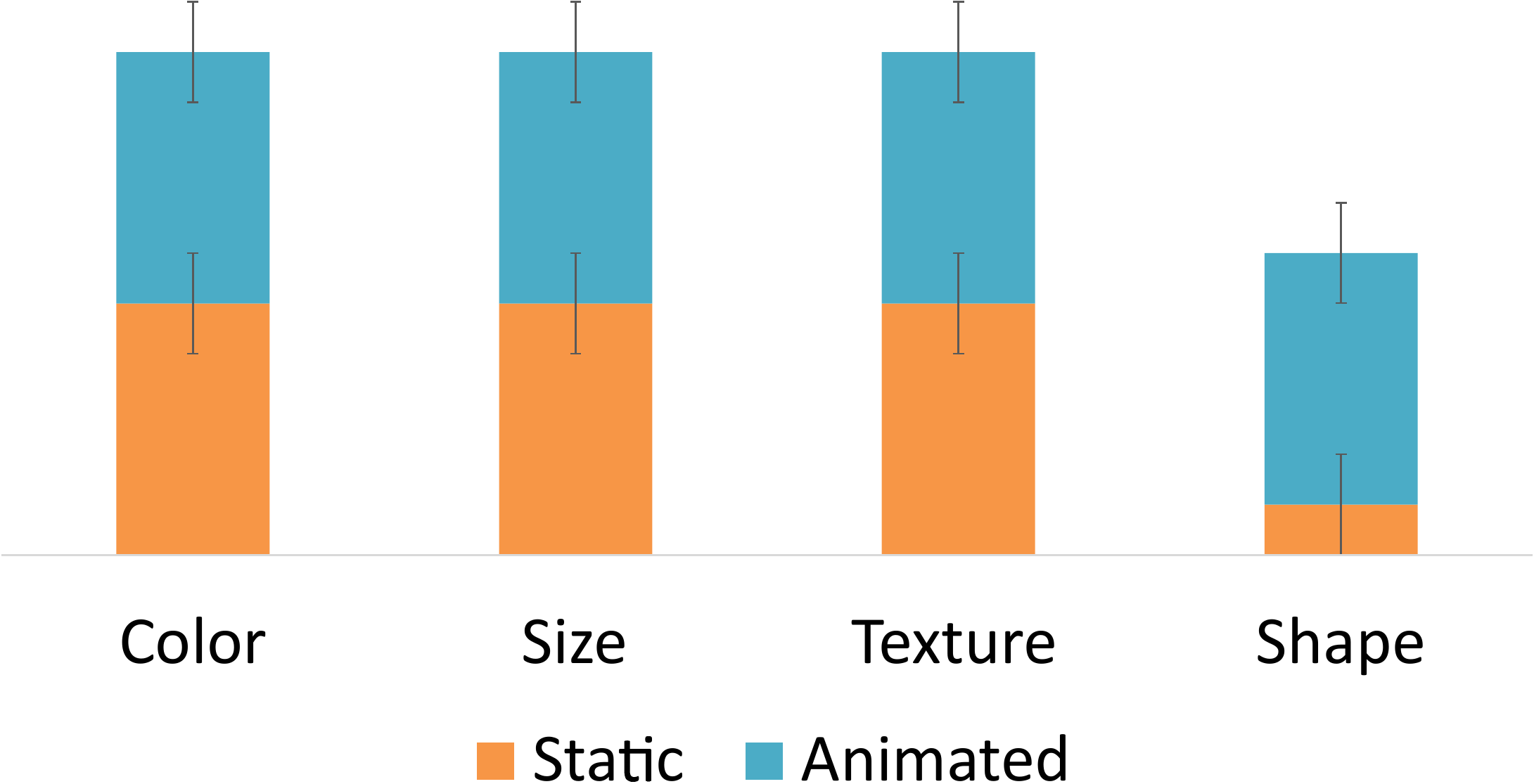}
                \caption{Randomly discarded object-feature \\ in scene description}
             \label{fig2:sfig6}
      \end{subfigure}%
      \begin{subfigure}[b]{0.33\textwidth}
                \centering
                \includegraphics[width=0.8\linewidth]{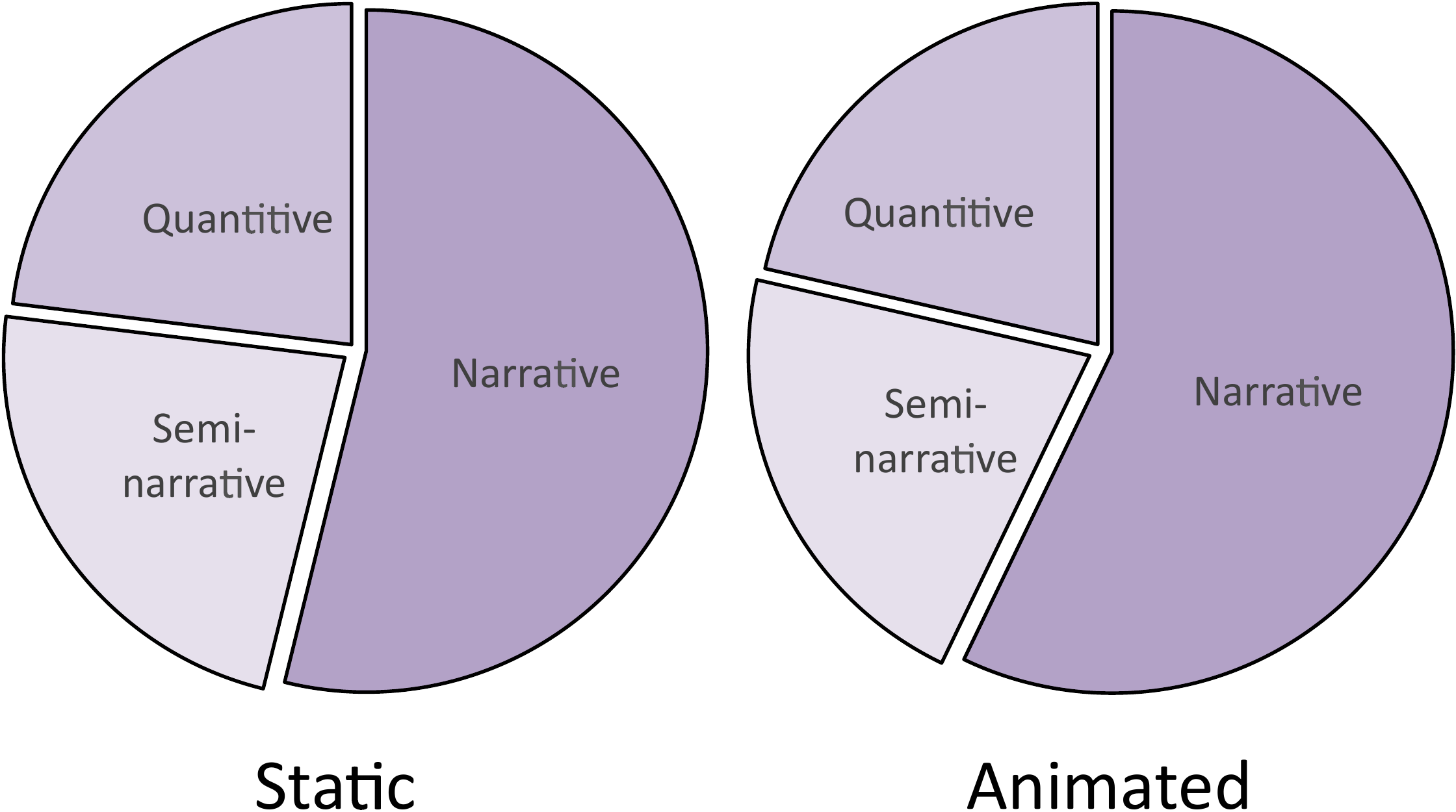}
                \caption{Distribution of description per category}
             \label{fig2:sfig7}
      \end{subfigure}
\caption{Different Distribution of our Dataset} \label{fig2}
\end{figure*}
%\anindya{At the beginning of each section write 1/2 lines on what is going to be discussed in this section.}
In this section, we describe in details the data generation process of our \textit{integrated Static and Animated Scene Generation (IScene) dataset.} We also show the analysis of our data distribution.

Our pipeline will take text descriptions as input and generate the corresponding 3D scenes. Hence, each data point in our dataset will comprise natural language scene description and corresponding attributes of the 3D scene. Under the current scope of our work, we consider three description families: \textit{Narrative}, \textit{Semi-Narrative} and \textit{Quantitative} for our free-form text description. %An example ground truth image with corresponding description of three families is shown in Figure \ref{fig:image_data}. 
A narrative description contains a detailed description of the scene. Here, information about each feature of each object type is mentioned explicitly. In figure \ref{fig:image_data}, we can see each object-feature of the cube \textit{(color: yellow, size: large, texture: metal)} is mentioned the narrative description. In practical cases, very detailed information of a scene might not always be available. To allow our pipeline perform with limited description, we consider semi-narrative descriptions. A semi-narrative description includes a subset of randomly chosen features of each object in a scene. In figure \ref{fig:image_data}, we can see that the size of the cube is not mentioned in the semi-narrative description. A quantitative description is a high-level abstraction of a scene where only the number of objects present in the scene are included. In figure \ref{fig:image_data}, we can see that there are two spheres present in the scene and corresponding quantitative description. We use templates to generate the corresponding descriptions based on the ground truth information. These \textit{description templates} can be extended further to generate even more types of descriptions. 
%(Mention clevr in footnote)

For an animated scene, we enhance ground truth information from CLEVR by inducing \textit{motion} following~\cite{cater}. Under the current scope of our work, we consider five simple atomic motions - \textit{Spin, Bounce, Shake, Move} and \textit{Rest}. We assign a motion with each scene object randomly. The effect of \textit{Rest} and \textit{Spin} are seemingly the same for cylinder and sphere, hence we do not consider \textit{Spin} for these two shapes in our data preparation. Figure~\ref{fig:sfig4} reports the distribution of the five motions in our dataset. Apart from \textit{Spin}, the distribution of the remaining four motions are even. Due to the restriction over the motion \textit{Spin}, it's occurrence is lower than the others. We generate motions as program templates which can be further extended to include a new variety of motions. The recorded animations are three seconds long by default. However, it can be tuned as well. The scene description for animated scenes is generated similarly to the static scenes under three families as described above. An example animated scene from our dataset with the corresponding narrative description is shown in Figure \ref{fig:video_data}.

\begin{table}[!h]
\centering
\resizebox{\linewidth}{!}{%
\begin{tabular}{|c|c|c|}
\hline
\begin{tabular}[c]{@{}c@{}}Description\\ Family\end{tabular} &
  \begin{tabular}[c]{@{}c@{}}Static Scene \\Description\end{tabular} &
  \begin{tabular}[c]{@{}c@{}}Animated Scene \\Description\end{tabular} \\ \hline
Narrative &
  \begin{tabular}[c]{@{}c@{}}Train: 4,90,000\\ condA: 1,05,000\\ condB: 1,05,000\end{tabular} &
  \begin{tabular}[c]{@{}c@{}}Train: 5,60,000\\ condA: 1,20,000\\ condB: 1,20,000\end{tabular} \\ \hline
\begin{tabular}[c]{@{}c@{}}Semi-Narrative\end{tabular} &
  \begin{tabular}[c]{@{}c@{}}Train: 2,10,000\\ condA: 45,000\\ condB: 45,000\end{tabular} &
  \begin{tabular}[c]{@{}c@{}}Train: 2,10,000\\ condA: 45,000\\ condB: 45,000\end{tabular} \\ \hline
\begin{tabular}[c]{@{}c@{}}Quantitative\end{tabular} &
  \begin{tabular}[c]{@{}c@{}}Train: 2,10,000\\ condA: 45,000\\ condB: 45,000\end{tabular} &
  \begin{tabular}[c]{@{}c@{}}Train: 2,10,000\\ condA: 45,000\\ condB: 45,000\end{tabular} \\ \hline
Total &
  13,00,000 &
  14,00,000 \\ \hline
\end{tabular}%
}
\caption{Total count of our dataset}
\label{tab:dataset}
\end{table}

There are multiple options and degrees of complicacy to describe a scene. For preparing our standard train-test-validation set, we prepare 13 and 14 \textit{description templates} for static and animated scenes, respectively that are further diversified by using synonyms. To divide our dataset in train-test-validation splits, we use the same data splits used in CLEVR, as it is an established benchmark. The total count of our dataset is shown in Table \ref{tab:dataset}. The distribution of each description family for static and animated scenes are shown in Figure \ref{fig2:sfig7}. We also provide some more details on our IScene dataset. Figure~\ref{fig2:sfig5},~\ref{fig:sfig1},~\ref{fig:sfig2},~\ref{fig:sfig3} reports the distribution of color, texture, size and shape, respectively. The distribution of each of these object-features in both static and animated scenes are even. Figure~\ref{fig2:sfig6} reports the distribution of missing object-feature information in semi-narrative and qualitative description. %Although the distribution of each object-feature is even in animated scenes, the distribution of shape is relatively lower for static scenes. This can be attributed as the inclusion of shape information in qualitative description while all other object-features are discarded. 
It is worth mentioning that all plausible biases are carefully discarded by weighted loss calculation during training the model~(see Section \ref{subsec:loss} for details).   

%Example of image and video with the corresponding output are shown in Figure~\ref{fig:data_example}.\footnote{More examples shown in the supplementary documents.}

\section{Our Pipeline}
\label{sec:model}
\begin{figure*}[!ht]
\centering
\includegraphics[width=\linewidth]{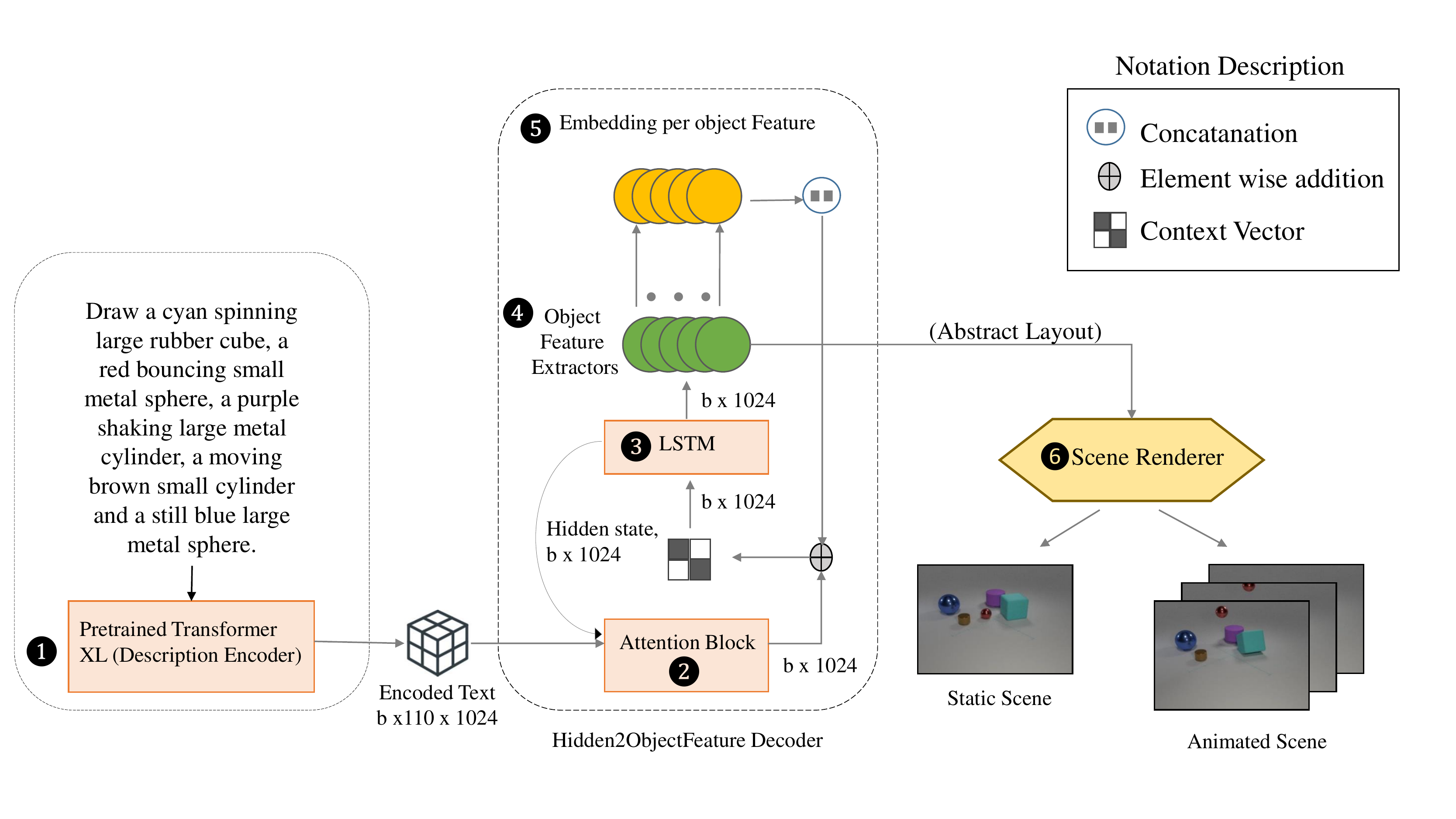}
\caption{The overall pipeline of our methodology being applied on a sample text description to generate the corresponding scene. (1) Text is encoded by TransformerXL (description encoder). (2) Attention is applied on the encoded text (from 1) and the previous hidden state of the LSTM (from 3). (3) LSTM takes as input the context vector (addition of the outputs from 5 and 2). (4) Object feature extractor generates the abstract layout of the scene and passes to 6. (5) Embedding per object feature embeds and concats the output from 4 and passes to 3. (6) Scene renderer generates the static or animated scenes from the abstract layout as specified by the user.}
\label{fig:architecture}
\end{figure*}

In this section, we describe different components of our pipeline. Our pipeline has three major components: Description Encoder (Section~\ref{subsec:pretrained}), Hidden2ObjectFeature Decoder (Section~\ref{subsec:Hidden2Feature}) and Scene Renderer (Section~\ref{subsec:script}). Figure~\ref{fig:architecture} shows the overall architecture of our pipeline.

\subsection{Description Encoder}
\label{subsec:pretrained}
To encode our scene description into a deep contextual vector, we use one of state-of-the-art architectures, TransformerXL~\cite{transformerxl}. We use the pretrained model of TransformerXL from HuggingFace Distribution~\cite{huggingface}. Transformers are proven to work well to capture the context and encode the information in a meaningful vector space. The motivation of including TransformerXL is to allow our pipeline capture the usefulness under low-resource constraints. The encoding is done according to the following equation:
\begin{equation}
H_0 = E(T_d)
\label{eqn:encoder}
\end{equation}
Where $T_d$ is the scene description containing $d$ words, $H_0$ $\in$ $\mathbb{R}^{d \times 1024}$ represents the encoded hidden vectors for all the input words and $E(.)$ is the pretrained TransformerXL model (as encoder).

\subsection{Hidden2ObjectFeature Decoder}
\label{subsec:Hidden2Feature}
We design a new multi-head decoder (referred as Hidden2ObjectFeature) which maps the encoded vector to object-feature distribution. Our decoder is specially architected to extract multiple object-features parallely and is capable to decode descriptions from different \textit{description families} efficiently. %The extracted object-features create an abstract layout which is then passed to the scene renderer to generate 3D scene. 

We use single-layer uni-directional LSTM~\cite{lstm} as our base decoder unit. We apply attention~\cite{attention} on the encoded hidden vectors (Section~\ref{subsec:pretrained}), $H_{0}$ and previous hidden state, $h_{t-1}$. Attention mechanism allows us the understand and capture the corresponding portions of input description which are more important while generating a specific output token. For example, intuitively, the words \textit{"a cyan spinning large rubber cube"} are most important while generating the cube as output in Figure~\ref{fig:architecture}. We will further analyze the output of our attention layer in Section~\ref{sec:result}. 

After applying attention on $H_{0}$ and $h_{t-1}$, we get the context vector,~$\hat{c}_t$. The addition of~$\hat{c}_t$ and input,~$i_{t}$ - is passed through the LSTM layer. Specifically,
\begin{gather}
a_t =  tanh(W_1 H_{0} + W_2 h_{t-1} + b_1)\\
\hat{a}_t = softmax(W_3 a_t + b_2)\\
\hat{c}_t = \hat{a}_t H_{0}\\
\label{eqn:LSTM}
y_t, h_t =  LSTM(\hat{c}_t + i_{t}, h_{t-1})
\end{gather}
Here $W_n, b_n$, and $h_n$ specify weight matrices, biases, and hidden vectors, respectively. $\hat{a}_t$ is the attention distribution over input vectors $H_0$ for generating the output $y_t$ as time-step $t$. The input $i_t$ is generated by the $N$ embedding layers of the decoder in the previous time step $t-1$ (to be elaborated later in Section \ref{multi_decoder_embedding})

%\begin{comment}
%% old
%The output,~$y_t$ from LSTM is passed through separate dense layers to map a particular object-feature. The motivation is to predict each object-feature separately because they are mutually independent. let $f^{[N]}$ be n object-features. The following formula calculates $f^{[N]}$:
%\begin{equation}
%f^{[N]} = \sum \limits_{i=1}^{n} softmax(W_i y_i + b_i)
%\label{eqn:feature}
%\end{equation}
%To further ensure equal emphasis of each object-feature, the input for the next state is obtained from the object-feature attributes of the current state. Specifically, let $i_{t+1}$, $\rho$ and $\varepsilon$ be the next input, dropout and embedding respectively. The following formula calculates $i_{t+1}$:
%\begin{equation}
%i_{t+1} = \rho * \varepsilon(f^{[N]})
%\end{equation}
%\end{comment}

The output, $y_t$ from LSTM is expected to be representing one object in the scene. $y_t$ is passed through $N$ separate dense layers to generate $N$ object-features, where $N$ is the number of object-features (depicted as \textit{Object Feature Extractors} in Figure \ref{fig:architecture}). The motivation is to predict each object-feature separately, because they are mutually independent. As we are using multiple dense layers parallely, our decoder is \textit{multi-headed} in nature.

Let $f_t=[f_{t,1}, f_{t,2},\dots, f_{t,N}]$ be $N$ object-features generated from $y_t$. Then, each $f_{t,i} ~\in ~f_t$ can be calculated using the Equation \ref{eqn:feature}.
\begin{equation}
f_{t,i} = softmax(W_{f_i}~y_t + b_{f_i})
\label{eqn:feature}
\end{equation}
%\anindya{Please give number to all equations.}
Here, $W_{f_i}$ and $b_{f_i}$ are the weight and bias of the dense layer that generates the $i-th$ object-feature $f_{t,i}$.

\subsection{Multi-head Decoder-Embedding}
\label{multi_decoder_embedding}
%\anindya{ Why not "Multiple Decoder Embedding"?}
To further ensure equal emphasis of each object-feature, the input for the next state is obtained from the object-features of the current state. For each of the object-features, there is one unique embedding layer $\epsilon_i(.)$, generating the object-feature-embedding vector $emb_{t,i}$ according to Equation \ref{eqn:object_feature_embedding}.

\begin{equation}
\label{eqn:object_feature_embedding}
emb_{t,i} = \epsilon_i(f_{t,i})
\end{equation}

The embedding vectors $[emb_{t,1}, emb_{t,2}, \dots, emb_{t,N}]$ are concatenated to generate the input vector $i_{t+1}$, which is then summed with the context vector generated by the attention mechanism (as shown in Equation \ref{eqn:LSTM}).

%. let $i_{t+1}$, $\rho$ and $\varepsilon$ be the next input, dropout and embedding respectively. The following formula calculates $i_{t+1}$:
%\begin{equation}
%i_{t+1} = \rho * \varepsilon(f^{[N]})
%\end{equation}

\subsection{Loss Function}
\label{subsec:loss}
Intuitively, each object-feature plays equal role for a successful scene generation. Hence, we calculate weighted negative log likelihood loss\footnote{\url{https://pytorch.org/docs/master/generated/torch.nn.NLLLoss.html}} for each object-feature. The weighting in considered to nullify any bias in the dataset.
% \begin{gather} 
% L(x, y) = \sum \limits_{i=1}^{n} \frac{-w_{y_{i}}*x_{i,y_{i}}} { \sum \limits_{j=1}^{n} w_{y_{j}}}
% \end{gather}

For calculating the loss, the rescaling weight given to each class is calculated according to the distribution of object-features of our dataset. This distribution is depicted in Figure \ref{fig2}.% For each object-feature, the weight is inversely proportional to the number of samples of a certain class (e.g. metal and rubber are two classes of texture).

\subsection{Object-Feature Accuracy Metric}
\label{subsec:feature_acc}
As each object-feature is equally important for a successful scene generation, during evaluation we check if each object-feature,~$f_{i}^{j}$ of each object matches with the ground truth information,~$t_{i}^{j}$. Hence, the Object-Feature Accuracy\footnote{Object-Feature Accuracy Calculation code \url{https://github.com/oaishi/3DScene_from_text/blob/master/scripts/evaluation_metric.py}}, $acc$, can be derived using Equation \ref{eqn:acc} and \ref{eqn:acc_helper}.

\begin{equation} 
acc = \frac{1}{l*n} \sum \limits_{i = 1}^{l}  \sum \limits_{j = 1}^n a_{i}^{j}
\label{eqn:acc}
\end{equation}
%https://tex.stackexchange.com/a/337352
\begin{equation}
  a_{i}^{j} =
    \begin{cases}
      0 & \text{if $f_{i}^{j} \neq t_{i}^{j}$}\\
      1 & \text{otherwise}
    \end{cases}       
\label{eqn:acc_helper}    
\end{equation}

Here, $l$ refer to the number of objects in the scene and $n$ is the number of object-features for each of the objects. %\anindya{Previously you used N to denote something else. So, better not to use n here. Try another variable.}
%\oaishi{sir both N refer to the same thing.}

\subsection{Scene Renderer}
\label{subsec:script}
To render the final 3D scene from the abstract layout output of decoder, we prepare a Blender scene renderer script. Our script takes the layout and perform all necessary post-processing which includes camera setup, light positioning, 3D model loading and animation setup.% An important functionality of scene renderer is to remove ambiguities such as collision among scene objects.     %the-work-process-of-scene-generator-is-shown-in-figure-?-with-the-overall-architecture

\section{Experiment}
\label{sec:experiment}
In this section, we discuss our experimental setup (Section~\ref{subsec:Setup}), evaluation metrics (Section~\ref{subsec:eval_met}), and hyperparameter settings (Section~\ref{subsec:hyper}).
\subsection{Experimental Setup}
\label{subsec:Setup}
To the best of our knowledge, there has not been any prior study on static and animated 3D scene generation from free-form text descriptions. Hence, We evaluate and compare between three variations of our proposed model. Our first model, $M_{static}$ is trained on static scene descriptions. The second model, $M_{animated}$ is trained on animated scene descriptions. The third and final model is an oracle model, $M_{full}$ that is trained on both static and animated scene descriptions. We prepare a separate training, validation and test set for each model. The overview of the data count used for each model configuration is used in Table~\ref{tab:splitted}. It is worth mentioning, each set of training, validation and test dataset contains non-overlapping data points selected from the whole dataset (Table~\ref{tab:dataset}).

\begin{table}[!h]
\centering
\resizebox{\linewidth}{!}{%
\begin{tabular}{|l|l|l|l|}
\hline
Model          & Train    & Validation                                                        & Test                                                               \\ \hline
$M_{static}$   & 1,00,000 & \begin{tabular}[c]{@{}l@{}}condA: 5000\\ condB: 5000\end{tabular} & \begin{tabular}[c]{@{}l@{}}condA: 6400\\ condA : 6400\end{tabular} \\ \hline
$M_{animated}$ & 1,00,000 & \begin{tabular}[c]{@{}l@{}}condA: 5000\\ condB: 5000\end{tabular} & \begin{tabular}[c]{@{}l@{}}condA: 6400\\ condB: 6400\end{tabular}  \\ \hline
$M_{full}$ &
  \begin{tabular}[c]{@{}l@{}}s: 50,000\\ a: 50,000\end{tabular} &
  \begin{tabular}[c]{@{}l@{}}condA: 2500(s) + 2500(a)\\ condB: 2500(s) + 2500(a)\end{tabular} &
  \begin{tabular}[c]{@{}l@{}}condA: 3200(s) + 3200(a)\\ condB: 3200(s) + 3200(a)\end{tabular} \\ \hline
\end{tabular}%
}
\caption{Train-Validation-Test Split (s and a denote static and animated description respectively)}
\label{tab:splitted}
\end{table}

\subsection{Evaluation Metric}
\label{subsec:eval_met}
We evaluate our pipeline on the following metrics.

\subsubsection{Object-Feature Accuracy}
As discussed in Section~\ref{subsec:feature_acc}, we evaluate if each object-feature is the same as the ground truth object-feature information. Object-feature accuracy will show how accurately our model (Section~\ref{sec:model}) can understand the natural language description.

\subsubsection{Structural Similarity Index (SSIM)}
Structural Similarity Index~\cite{ssim} is a standard metric for predicting the perceived quality of images and videos~\cite{storygan}. After the final scene is rendered (Section~\ref{subsec:script}), we use SSIM to compare the generated image (static scene) and video (animated scene) with the ground truth. The result will show how accurately our pipeline can generate scenes with respect to ground truth as well as understand the natural language description.

% \begin{figure*}[!h]
% \includegraphics[width=\linewidth]{images/joint_dataset_marked.eps}
% \caption{Successful example of generated and ground truth static scene with corresponding \textbf{narrative} description (left)~\& generated and ground truth animated scene with corresponding \textbf{semi-narrative} description (right) during inference}
% \label{fig:data_example}
% \end{figure*}

\subsection{Hyperparameter Settings}
\label{subsec:hyper}
We did an extensive search over the possible space of hyperparameters using the validation set (discussed in Section \ref{subsec:Setup}) to select learning rate (0.0001 to 0.1), dropout rate (0 to 0.8), LSTM hidden dimension (128 to 2048), teacher forcing ratio (0\% to 80\%). The best set of hyperparameters is mentioned below which we use throughout our final experiments:

%\anindya{Can you give justification for at least some parameteres like an Ablation study? This is important.}

\begin{itemize}
    \item Learning Rate = $0.01$
    \item Dropout Rate = $0.1$
    \item Teacher Forcing Ratio = $50\%$
    \item LSTM Input Dimension = Dimension of the Context vector, $\hat{c}_t$ (discussed in Section \ref{sec:model}, Equation \ref{eqn:LSTM}) = $1024$ \item LSTM Hidden Dimension = Dimension of Final Dense Layer Input, $y_t$ (discussed in Section \ref{sec:model}, Equation \ref{eqn:feature}) = $1024$
    %\item Dense Layer Input Dimension = $b \times 1024$ 
    %\item Dense Layer Output Dimension = ${b \times f_{t}}$ (following equation \ref{eqn:feature})
    
\end{itemize}

We ran all the experiments and analyses with a batch size of 520 on the same machine with Intel core i7-7700 CPU (4 cores), 16GB RAM, NVIDIA GeForce GTX 1070 GPU (8GB memory). During training, all our models reach convergence within 120 epochs and take approximately 78 hours.

\section{Result and Discussion}
\label{sec:result}
%\anindya{There should be 1/2 line introductory discussion about what will be presented in this Section.}

In this section, We present the results obtained by our models. We also provide several analysis over the generated scenes and model performance.  

\begin{figure}[!ht]
\centering
\includegraphics[width=0.9\linewidth]{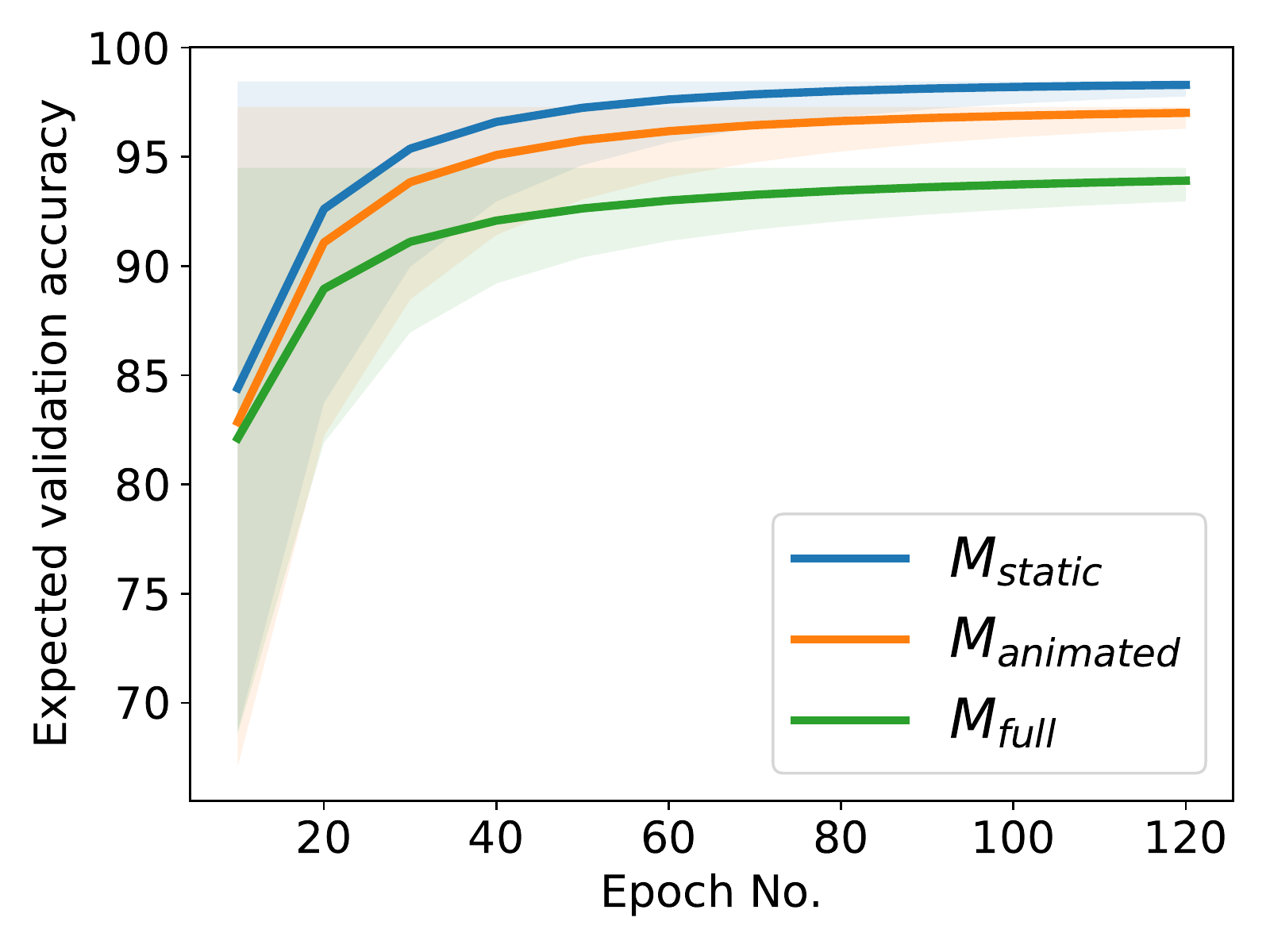}
\caption{The expected validation performance for $M_{static},~M_{animated},~M_{full}$~\cite{showyourwork}}
\label{fig:valdation_curve}
\end{figure}

\begin{figure*}[ht]
    \begin{subfigure}[t]{0.29\textwidth}
        \centering
          \includegraphics[width=0.9\linewidth]{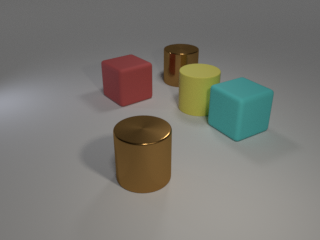}
          \caption{\textbf{Description:} Draw a large yellow colored cylinder of matte texture, a large cyan colored cube of matte texture, 
          a large brown colored cylinder of shiny texture, a large red colored cube of matte texture and a 
          large brown colored cylinder of shiny texture.}
          %\label{fig:image_data}
    \end{subfigure}%
    \hspace{2mm}
    \begin{subfigure}[t]{0.29\textwidth}
        \centering
          \includegraphics[width=0.9\linewidth]{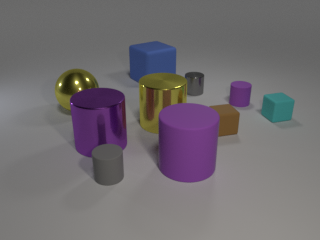}
          \caption{\textbf{Description:} Draw a small brown matte object, a small cyan matte object, a large matte cylinder, a large yellow shiny object, a small matte sphere, a small gray cylinder, a blue shiny cylinder, a large shiny cylinder, a gray matte cylinder and a large yellow shiny object.}
          %\label{fig:video_data}
    \end{subfigure}
    \hspace{2mm}
    \begin{subfigure}[t]{0.29\textwidth}
        \centering
          \includegraphics[width=0.9\linewidth]{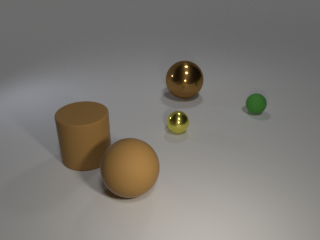}
          \caption{\textbf{Description:} There are four spheres and one cylinder.}
          %\label{fig:video_data}
    \end{subfigure}
\vspace{3mm}
\caption{Few examples of generated static scene using our pipeline with corresponding input descriptions.} 
\label{fig:static_exam}
\end{figure*}

We train each of our models for 120 epochs until convergence with a validation step after every 10 epochs. As Figure~\ref{fig:valdation_curve} reports, the expected performance stabilizes and the error decreases significantly after 120 epochs. At this point, we report our final accuracy over the metrics discussed in Table \ref{tab:accuracy} in Section~\ref{subsec:eval_met}. Here we show evaluation on both known and novel combinations of object feature (referred as ConditionA and ConditionB in section \ref{sec:task} respectively). We show some of the generated static and animated scenes during testing in Figure \ref{fig:static_exam} and \ref{fig:anim_exam} respectively.

\begin{table}[!ht]
\centering
\resizebox{\linewidth}{!}{%
\begin{tabular}{|c|c|c|c|c|}
\hline
\multirow{2}{*}{Model} &
  \multicolumn{2}{c|}{ConditionA (Known combination)} &
  \multicolumn{2}{c|}{ConditionB (Novel combination)} \\ \cline{2-5} 
 &
  \begin{tabular}[c]{@{}c@{}}Object Feature \\ Accuracy\end{tabular} &
  SSIM &
  \begin{tabular}[c]{@{}c@{}}Object Feature\\ Accuracy\end{tabular} &
  SSIM \\ \hline
$M_{static}$   & 98.427 & 0.801 & 94.270 & 0.809 \\ \hline
$M_{animated}$ & 97.482 & 0.906 & 93.234 & 0.857 \\ \hline
$M_{full}$ &
  94.910 &
  \begin{tabular}[c]{@{}c@{}}static: 0.812\\ animated: 0.849\end{tabular} &
  90.040 &
  \begin{tabular}[c]{@{}c@{}}static: 0.813\\ animated: 0.888\end{tabular} \\ \hline
\end{tabular}%
}
\caption{Accuracy and SSIM of our models on test set}
\label{tab:accuracy}
\end{table}

The results show that our pipeline performs consistently for all of $M_{static},~M_{animated},~M_{full}$. The slight decrease in object-feature accuracy of $M_{animated}$ from $M_{static}$ is due to the inclusion of motion which enforces the model to learn one extra object-feature (motion) than $M_{static}$. Similarly, the decrease in object-feature accuracy of $M_{full}$ is due to the extra constraints over the model. However, we see that the relative difference between these configurations is negligible. This proves the effectiveness of our proposed pipeline for both static and animated scenes. The accuracy also shows that the model performs almost similarly on joint (both static and animated scene using the same model) generation and disjoint (only static or animated scene using a single model) generation. This proves the robustness of our architecture to generate static as well animated scenes using the same configuration.
%This provides the answer to RQ2 (Section \ref{sec:Introduction}). \anindya{Answer to RQ1 should come first or the first one should be named RQ1.}

% https://tex.stackexchange.com/a/125294

\begin{figure*}[!ht]
\centering
\includegraphics[width=\linewidth]{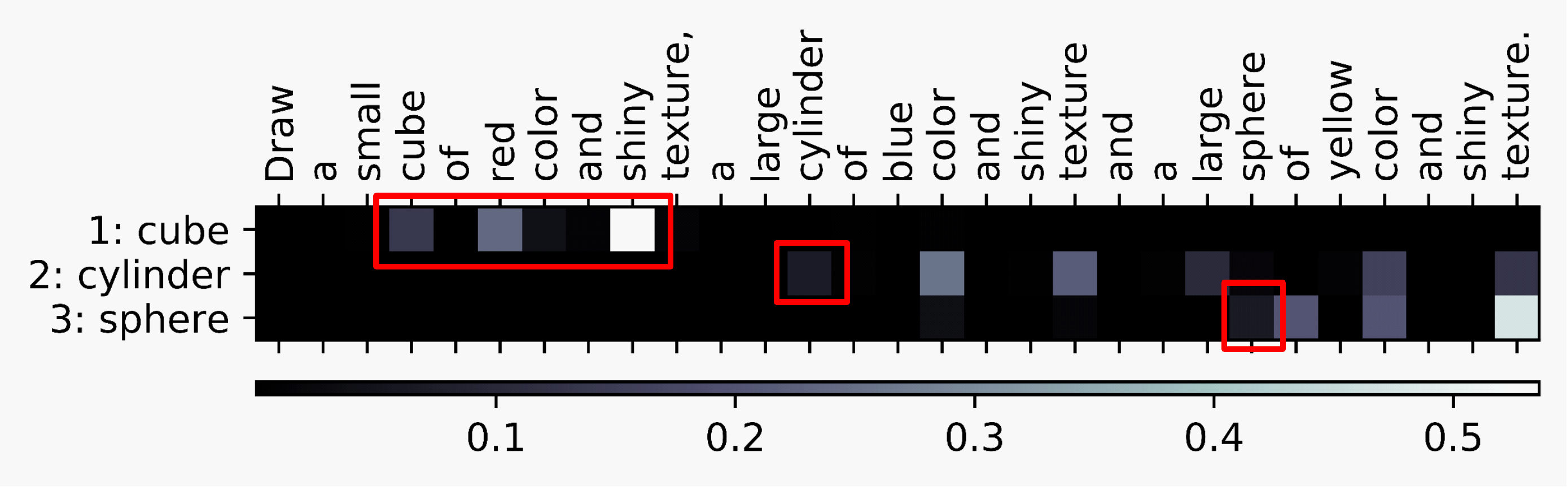}
\caption{Attention map for a successful test sample generated by our pipeline. Scene description and predicted outputs are shown on X and Y axes, respectively.} %Words related to different attributes are given more emphasis to generate an object.}
\label{fig:attn_map}
\end{figure*}

\begin{figure*}[ht]
    \begin{subfigure}[t]{\textwidth}
        \centering
          \includegraphics[width=\linewidth]{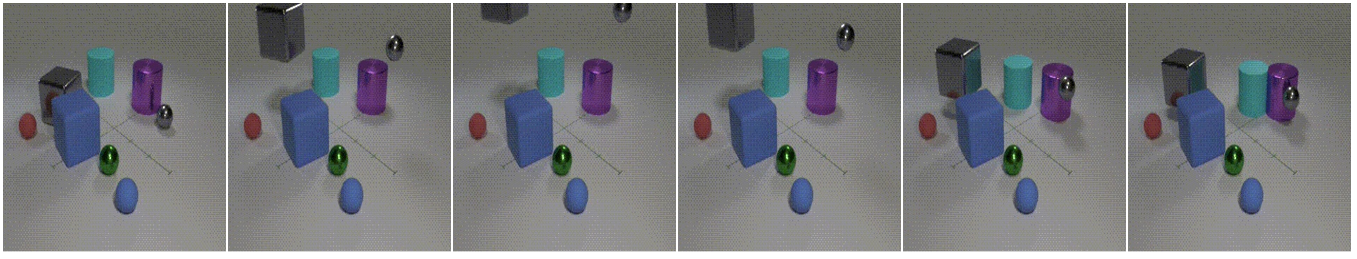}
          \caption{\textbf{Description:} Draw a green metal small sphere, a purple metal large cylinder, a red rubber small sphere, a gray metal small sphere, a blue rubber large cube, a cyan rubber large cylinder, a gray metal large cube and a blue rubber small sphere.}
          %\label{fig:image_data}
    \end{subfigure}%
    \vspace{3mm}
    \begin{subfigure}[t]{\textwidth}
        \centering
          \includegraphics[width=\linewidth]{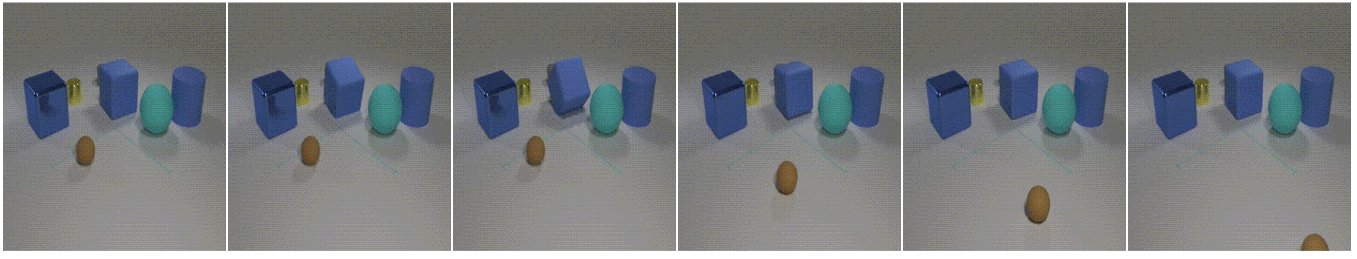}
          \caption{\textbf{Description:} A rocking cyan matte sphere, a small rocking gray matte object, a small rocking shiny cylinder, a large rocking blue matte object, a spinning blue matte cube, a small moving brown sphere and a rocking blue shiny cube. (One of the objects is occluded in the figure)}
          %\label{fig:video_data}
    \end{subfigure}%
    \vspace{3mm}
    \begin{subfigure}[t]{\textwidth}
        \centering
          \includegraphics[width=\linewidth]{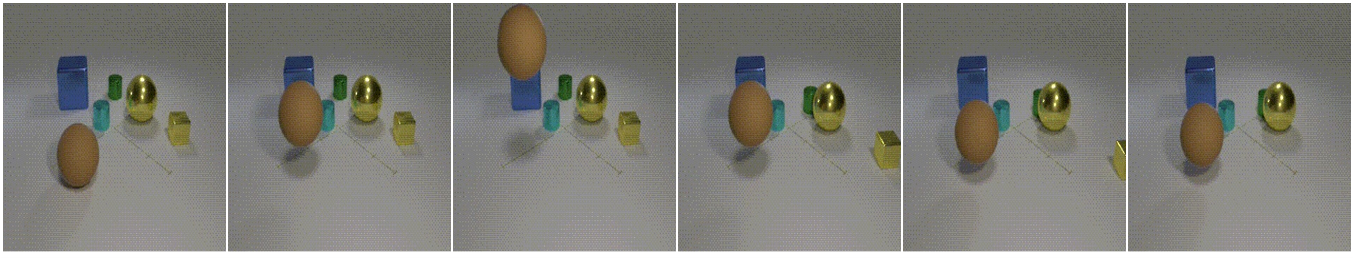}
          \caption{\textbf{Description:} There are a yellow moving metal large sphere, a cyan still metal small cylinder, a yellow moving metal small cube, a green moving metal small cylinder, a blue rocking metal large cube and a brown bouncing rubber large sphere.}
          %\label{fig:video_data}
    \end{subfigure}
\vspace{3mm}
\caption{Few examples of generated animated scene using our pipeline with corresponding input descriptions.} 
\label{fig:anim_exam}
\end{figure*}

The performance of each model on known and novel combination is also noteworthy. The small margin of accuracy between these combinations shows our pipeline has a very small bias on data pattern and works efficiently for novel description as well. 
%This provides the answer to RQ1 (Section \ref{sec:Introduction}).
Overall, the higher accuracy strengthens the applicability of our system for text-to-scene generation.

\subsection{Case Study}
\begin{figure*}[!ht]
\vspace{6mm}
\includegraphics[width=\linewidth]{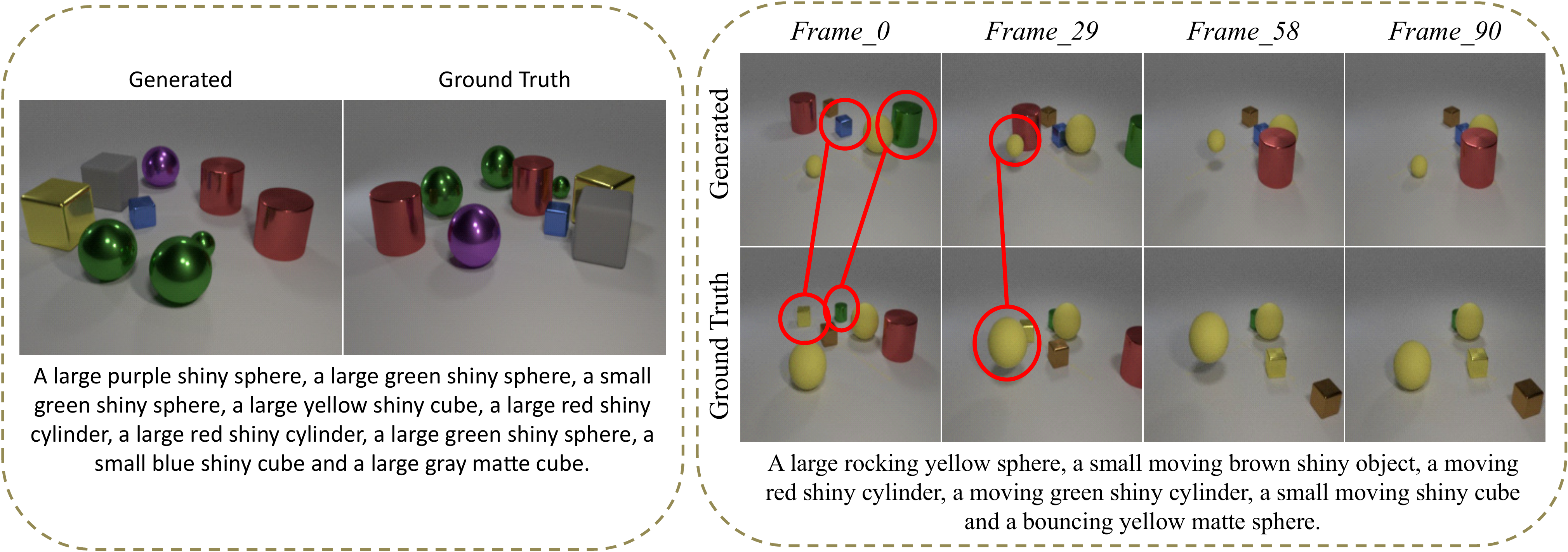}
\caption{Case Study using two successful example of generated and ground truth static scene with corresponding \textbf{narrative} description (left)~\& generated and ground truth animated scene with corresponding \textbf{semi-narrative} description (right) during inference.}
\label{fig:data_example}
\end{figure*}

In this section, we analyze two particular outputs from our test set~\footnote{More examples are available at \url{https://github.com/oaishi/3DScene_from_text}.} generated by our pipeline as shown in Figure~\ref{fig:data_example}. 

On the left, we can see the generated and ground truth static scenes with corresponding \textit{narrative} scene description. Here we observe that every feature of an object, i.e., color, shape, size, and texture matches with the ground truth for all nine scene objects. %However, the relative positioning differs from the ground truth. 

On the right, we can see the generated and ground truth animated scenes with corresponding \textit{semi-narrative} scene description. As we can see, every motion of all the six objects matches with the ground truth and input description. However, \textit{`a small moving shiny cube'} has been mapped to a \textit{`a small \textbf{blue} moving shiny cube'} whereas ground truth scene is \textit{`a small \textbf{yellow} moving shiny cube'}. Our pipeline successfully maps the texture, shape, and size mentioned for this object. As the color of the cube was not mentioned in the input, it was impossible to infer the ground truth information, and hence the generated animated scene could not reflect it.

\subsection{Result Analysis}
%We illustrate insights from our model performance. 
In this section, We analyze if our model captures the scene context from description correctly. For this, we focus on the output of the attention layer discussed in Section~\ref{subsec:Hidden2Feature}. Figure \ref{fig:attn_map} shows the attention map of a successful test example during inference. As the value gets closer to 1 and as the color goes closer to white, it depicts that the word is given more importance (higher attention weight) by our model. As expected, \textit{cube, red, color, shiny} are more important while generating the first object (i.e., `cube'). Similarly, we can see that the words \textit{cylinder} and \textit{sphere} are more important while generating the second and third objects, i. e., `cylinder' and `sphere', respectively.% Hence, despite some false positive values (unmarked words in the images), our pipeline correctly captures the input context.
%For example, the last word `texture' is not much important for the second object  `cylinder'.

\begin{figure*}[!ht]
\centering
\includegraphics[width=0.8\linewidth]{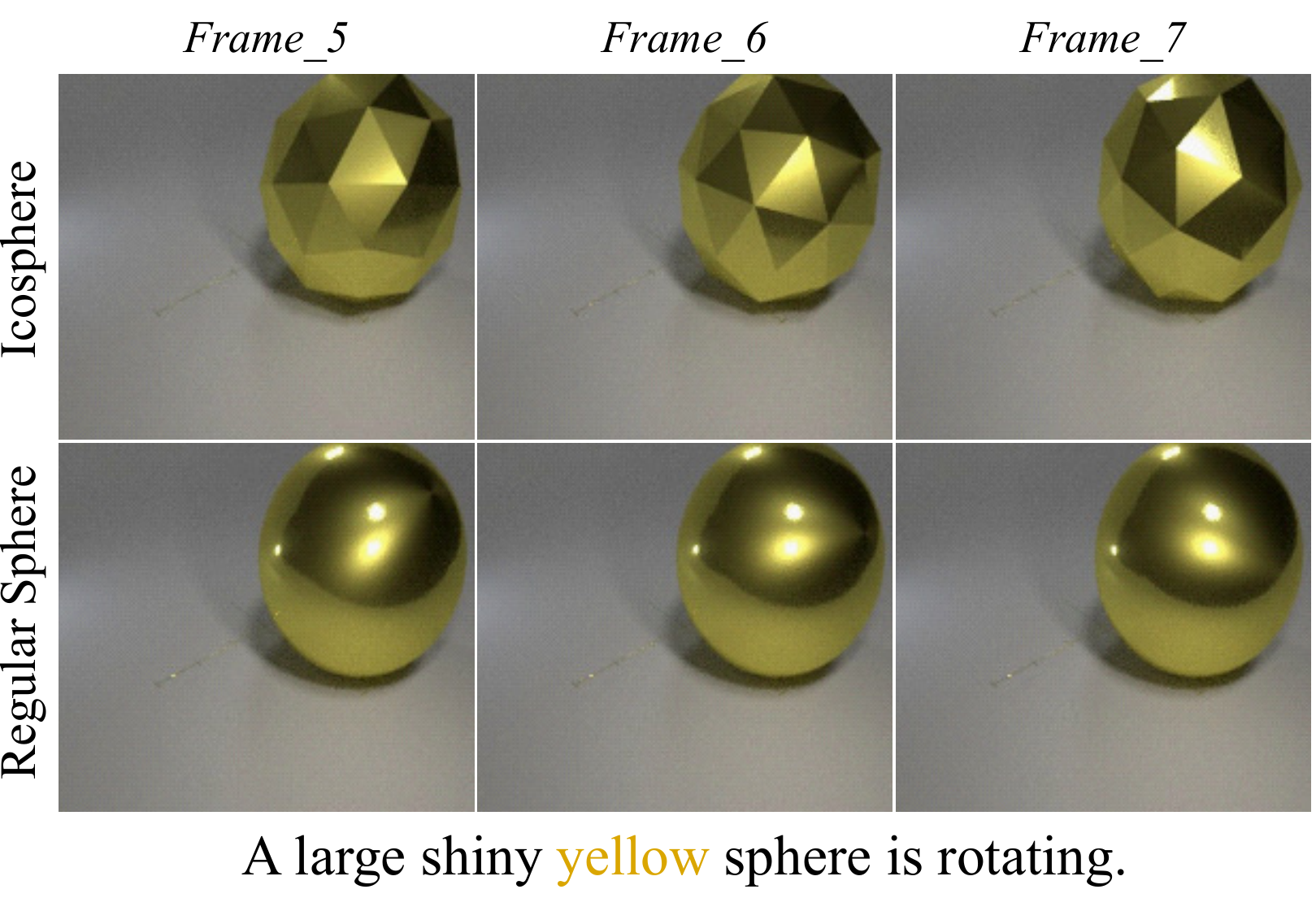}
\caption{The same description is being mapped to two different models of sphere without hampering the performance. Our pipeline thus allows us to use different 3D models even during inference}
\label{fig:adaptability}
\end{figure*}

\subsection{Editability} 
In practical life, we will often find the need of new or modified 3D models. For example, each movie has a unique set of characters. Hence, we want our system to be able to work on different variations of 3D models. Hence, we show the editability of our pipeline which will enable our system to process novel models unseen at training time. %Our pipeline enables us to change and control the scene objects. 
Figure \ref{fig:adaptability} reports a successful datapoint in inference. Although the descriptions of both the scenes are the same, the first scene includes an icosphere which represents a sphere in low-poly graphics. The second scene includes the regular smooth sphere used in the literature. Thus, using the same description and same training parameters, our pipeline allows us to control and change the settings, models, and features if needed.  %This is practically more useful.
%\anindya{Explain in which usecases this is useful.} 
%\oaishi{addressed, sir.} 

% \subsection{Failure Case}
% A relevant example

\section{Conclusion and Future Work}
\label{sec:conclusion}
%We introduce a novel pipeline of holistic static and animated 3D scene generation from free-form text descriptions. To do this, we deploy state-of-the-art natural language understanding architecture, a new decoder and observe a promising result. We further show various analysis and strengths of our pipeline. Our proposed method can generate static as well as animated scene with a longer coverage of scene descriptions. In future, we wish to explore advanced Graph Network modules, for more complex scene description including relational description to handle relative positioning.% We modify CLEVR to generate our dataset, IScene. In future, we would deploy our pipeline on a more practical real-life environment, for example, indoor 3D scene synthesis. 

In this work, we have built a pipeline that generates static as well as animated 3D scenes from different free-form scene descriptions. For our experiment, we have prepared a large synthetic dataset which consists of 13,00,000 and 14,00,000 unique static and animated scene descriptions, respectively. We achieve upto 98.427\% accuracy on test dataset in detecting the 3D objects features successfully. We also showed how our pipeline can adapt to new and unseen 3D models at the test phase. We believe that our approach would assist greatly in 3D production. 

In future, we wish to further extend our dataset for accommodating more complex scene descriptions and explore advanced Graph Networks. We also wish to explore the applicability of our system in the context of real-world. A possible application can arise immediately in children education. It can help children in learning basic mathematics using visual component. For example, if we generate one scene with \textit{four} spheres and another scene with \textit{two} spheres - we can teach them that there are \textit{two} spheres less in the second scene than the first one. Similarly, we can use our system to teach children in other subjects besides mathematics.  

\bibliographystyle{acl_natbib}
\bibliography{main.bbl}

\end{document}